%% file: main.tex
\documentclass{article}

\usepackage[preprint]{neurips_2024}

\usepackage[utf8]{inputenc} 
\usepackage[T1]{fontenc}    
\usepackage{hyperref}       
\usepackage{url}            
\usepackage{booktabs}       
\usepackage{amsfonts}       
\usepackage{nicefrac}       
\usepackage{microtype}      
\usepackage{xcolor}         
\usepackage{epsfig,verbatim}
\usepackage{amsmath,amssymb}
\usepackage{amsthm,amsfonts}
\usepackage{color}
\usepackage{float}
\usepackage{diagbox}
\usepackage{multirow}
\usepackage{natbib}
\usepackage{booktabs}
\usepackage{algorithm}
\usepackage{verbatim}
\usepackage{enumitem}
\usepackage{algorithmic}
\usepackage{subcaption} 
\usepackage{caption}
\usepackage{graphicx}

\input{math_commands}

\title{
Post-hoc Interpretability Illumination for\\ Scientific Interaction Discovery
}

\author{%
  Ling Zhang \\
  Adobe Inc. \\
  San Jose, CA, 95110 \\
  \texttt{zlaajj@gmail.com} \\
  \And
   Zhichao Hou \\
   North Carolina State University \\
   Raleigh, NC, 27695 \\
   \texttt{zhou4@ncsu.edu} \\
    \And
   Tingxiang Ji \\
   North Carolina State University \\
   Raleigh, NC, 27695 \\
   \texttt{tji2@ncsu.edu} \\
   \And
   Yuanyuan Xu \\
   University of Delaware \\
   Newark, DE, 19716 \\
   \texttt{yyx@udel.edu} \\
   \And
   Runze Li \\
   Pennsylvania State University \\
   State College, PA, 16801 \\
   \texttt{rzli@psu.edu} \\
}

\begin{document}

\maketitle

\input{section/abs}

\input{section/intro}

\input{section/relate}

\input{section/method}

\input{section/exp}

\input{section/conclusion}

\section*{Broader Impact}

This paper proposes a new variable selection and interaction identification algorithm for high-dimensional data, which
have frequently been collected in many research areas including genomics,
biomedical imaging, functional magnetic resonance imaging, tomography, tumor
classifications and finance. Thus, the proposed procedure is expected
to benefit a broad range of scientists and researchers in various fields.

\bibliography{kfs}
\bibliographystyle{chicago}

\appendix

\input{section/app}

\end{document}

%% file: math_commands.tex
\usepackage{amsmath,amsfonts,bm}

\def\eqref#1{(\ref{#1})}

\def\1{\bm{1}}

\newcommand{\vw}{{\mathbf{w}}}
\newcommand{\vx}{{\mathbf{x}}}

\newcommand{\vB}{{\mathbf{B}}}

\newcommand{\vW}{{\mathbf{W}}}

\def\vw{{\bm{w}}}
\def\vx{{\bm{x}}}

\DeclareMathAlphabet{\mathsfit}{\encodingdefault}{\sfdefault}{m}{sl}
\SetMathAlphabet{\mathsfit}{bold}{\encodingdefault}{\sfdefault}{bx}{n}



%% file: section/abs.tex
\begin{abstract}

Model interpretability and explainability have garnered substantial attention in recent years, particularly in decision-making applications. However, existing interpretability tools often fall short in delivering satisfactory performance due to limited capabilities or efficiency issues. To address these challenges, we propose a novel post-hoc method: Iterative Kings' Forests (iKF), designed to uncover complex multi-order interactions among variables. iKF iteratively selects the next most important variable, the "King", and constructs King's Forests by placing it at the root node of each tree to identify variables that interact with the "King". It then generates ranked short lists of important variables and interactions of varying orders. Additionally, iKF provides inference metrics to analyze the patterns of the selected interactions and classify them into one of three interaction types: Accompanied Interaction, Synergistic Interaction, and Hierarchical Interaction. Extensive experiments demonstrate the strong interpretive power of our proposed iKF, highlighting its great potential for explainable modeling and scientific discovery across diverse scientific fields.

\end{abstract}

%% file: section/intro.tex
\section{Introduction}

In recent years, numerous sophisticated models have emerged, aiming to improve accuracy, efficiency, and robustness across various domains~\citep{lin2022survey, sze2017efficient, xu2020adversarial, hou2024adversarial}. However, model interpretability and explainability have been relatively overlooked and have not kept pace with the rapid development of emerging model architectures. Given the importance of interpretability in decision-making contexts, such as in medical diagnostics and epistasis interaction detection, it is imperative to design effective techniques to enhance explainability and to understand how models process information and make decisions.

To address this research gap, many researchers have explored ways to improve interpretability from various perspectives. On one hand, numerous methods have been developed to select important variables. In statistics, methods including regularization methods~\citep{tibshirani1996regression,fan2001variable,zou2005regularization,zou2008one}, feature screening methods ~\citep{fan2008sure,li2012feature,yang2019feature} and random forest-based techniques~\citep{breiman2001random, diaz2006gene, amaratunga2008enriched} have shown substantial capability for variable selection, but these methods often highly rely on model assumptions, limiting their application to a wider range of scenarios.
Furthermore, perturbation-based approaches~\citep{ma2013visualizing, zhou2015predicting, zintgraf2017visualizing} and attribution techniques~\citep{ribeiro2016explaining, shrikumar2017learning, ancona2017towards, lundberg2017unified} have facilitated interpretability in black-box neural networks. However, while effective at identifying main variable effects, these methods typically fail to capture interactions among variables.

Beyond individual variable effects, researchers have sought to capture variable interactions. Statistical methods~\citep{yuan2006model, simon2013sparse, shah2014random, basu2018iterative} are commonly used to detect interaction effects in linear and random forest-based models. In bioinformatics, for example, studies such as~\citep{elmarakeby2021biologically, zhao2021deepomix, hou2023pathexpsurv} have developed biologically-informed neural networks using pathway databases to uncover potential group effects beyond existing knowledge. 

However, most of these methods suffer from one or more of the following limitations: (1) they often fail to capture complex multi-order interactions, especially hierarchical interactions that are critical in bioinformatics; (2) they may be tied to specific model architectures or heavily reliant on domain knowledge, limiting their broader applicability; and (3) incorporating interactions during training can hinder generalization to large-scale pre-trained models.

To overcome these limitations, we propose a novel approach, the \textit{Iterative Kings' Forests (iKF)}, a post-hoc interpretability tool designed to select complex interactions and identify their type. Our contributions are as follows:

\begin{itemize}[left=0pt]
    \item iKF is capable of identifying key variables and complex interactions, and can be applied to any model, including large-scale models, without the need for retraining.
    \item iKF can identify fine-grained interaction structures, including three interaction types: Accompanied Interactions, Synergistic Interactions, and Hierarchical Interactions, as described in section 3.3.
    \item Extensive experiments and a real biological scientific re-discovery highlight the potential of iKF in scientific discovery, as detailed in sections 4 and 5.
\end{itemize}

The remainder of this paper is organized as follows. Section 2 provides an overview of related work in model interpretability. Section 3 introduces the preliminary concepts related to tree structures and presents our iKF method. Sections 4 and 5 evaluate iKF's performance in scientific discovery through diverse experimental scenarios and a biological case study, respectively. Finally, Section 6 offers conclusions and discussions.

%% file: section/relate.tex
\section{Related Works}

In this section, we present an overview of current methods aimed at improving model interpretability by identifying key variables and understanding their intrinsic interactions.

Various approaches have been developed to select important variables. Regularization methods~\citep{tibshirani1996regression,fan2001variable,zou2005regularization,zou2008one} use regularization terms for variable selection, especially in high-dimensional data contexts. Variable screening methods ~\citep{fan2008sure,li2012feature,yang2019feature} formulate computationally efficient procedures for ranking and screening variables, making them well-suited for ultra-high-dimension scenarios. Methods based on random forests~\citep{breiman2001random,diaz2006gene,amaratunga2008enriched} select important variable by calculating importance scores or iteratively dropping variables. Perturbation-based approaches~\citep{ma2013visualizing,zhou2015predicting,zintgraf2017visualizing} introduce data input perturbations and measure variable importance through changes in hidden or output layers. Additionally, attribution techniques, including LIME~\citep{ribeiro2016explaining}, DeepLIFT~\citep{shrikumar2017learning}, DeepExplain~\citep{ancona2017towards}, and SHAP~\citep{lundberg2017unified}, backpropagate importance signals from output neurons to inputs, facilitating the interpretability of black-box neural networks.

Beyond identifying individual variable effects, researchers have also sought to capture variable interactions. Statistical methods~\citep{yuan2006model, simon2013sparse, shah2016modelling} explore the collective effects of variable groups, while random forest-based approaches~\citep{shah2014random,basu2018iterative} are widely used to detect interaction effects. Variable interactions are especially significant in bioinformatics, where \citep{elmarakeby2021biologically, zhao2021deepomix}
construct biologically-informed neural networks with existing pathway databases, considering gene group effects within signal pathways or functional modules. Building on this foundation, \citep{hou2023pathexpsurv} introduce PathExpSurv, a novel method for uncovering potential group effects beyond existing database knowledge. However, these methods typically lack sophisticated architectures capable of capturing fine-grained interaction structures. They also fail to detect the nested nature of interactions in some special scenarios including biological gene expression regulation or dictionary-like log data~\citep{levine2010transcriptional, hou2024hlogformer}, which is crucial but largely overlooked. Additionally, many methods are tied to specific model architectures or rely heavily on domain knowledge, limiting their broader applicability. Moreover, capturing interactions during training will hinder generalization to large-scale models.

To address these limitations, we propose a novel post-hoc interpretability tool, the Iterative Kings' Forests, to identify complex interactions, including high-order and sequentially ordered interactions.

%% file: section/method.tex
\section{Iterative Kings' Forests}

In this section, we will first introduce the basics of tree structures and their natural rationale for modeling variable interactions. Then, we will highlight the bottlenecks of existing tree-based methods, which motivate the development of our proposed iterative Kings' Forests (iKF).

\textbf{Notation.} Denote $p$-dim input variables as $\vx=\left(x_1,...,x_p\right)^\top\in\mathbb{R}^p$, 
$D$ is the maximum depth of any tree, $K$ is the number of the leaf nodes. We denote the indices of  $D$ chosen variables along a root to leaf path over entire $p$ variables as $(i_1,i_2,...,i_D)\subseteq [p]$.

\subsection{Properties of Tree Structure and PVIM}

Recent researches suggest that random forest algorithms perform well in identifying interactions~\citep{shah2014random,basu2018iterative}, 
largely due to the inherent properties of the tree structure:
\begin{itemize}[left=0pt]
\item \emph{Hierarchical Structure:} In the tree construction process, the split at every node,
except for the root node, depends on the previous splits made by its ancestor nodes. This ancestor-descendant hierarchical relationship makes the tree a natural structure for modeling interactions.
\item \emph{Information Gain Search Engine:} For each node, the greedy algorithm used to search for the optimal split selects the variable that results in the largest impurity decrease, which could be due to a main effect, an interaction, or even random noise.
\end{itemize}
Together, these two properties enable tree structure to effectively model both main effects and interactions simultaneously, making it a comprehensive structure for interaction discovery.

\textbf{Preliminary on tree structure.}
  Consider the leaf node $k\in[K]$, there is a corresponding indicator function and a $D$-depth path from the root node all the way to the leaf node $k$. We denote the set of samples which falls into this path as $\vB_k=\{\vx: x_{i_{1}}\in B_{k,1},\ldots,x_{i_{D}}\in
B_{k,D}\}$, where $B_{k,d}$ is an interval with form
$(-\infty, s_{k,d})$ or $[s_{k,d}, \infty)$, and $s_{k,d}$ is chosen to
split the node at depth $d$ in path $k$. Therefore, 
a tree $f(\vx)$  can be represented as a linear combination of $K$ indicator functions as: 
\begin{equation}
f(\vx)=\sum_{k=1}^{K}a_{k}I(\vx\in
\mathbf{B}_{k})=\sum_{k=1}^{K}a_kI(x_{i_{1}}\in B_{k,1},\ldots,x_{i_{D}}\in
B_{k,D}),
\end{equation}


Due to the two properties of tree structure, the
selected $D$ variables
may include main effects, interactions, and other randomly selected variables. To
identify main effects and interactions effectively, we
want to exclude unrelated variables, put variables from an interaction in the
same path, and identify the interactions’ order as much as possible. The
proposed iKF algorithm provides a framework for these purposes.

\textbf{Permutation variable importance measure (PVIM).} In tree structure, PVIM~\citep{breiman2001random}  is proposed to conduct variable selection through ranking a measure of variables' importance:
\begin{equation}
\text{PVIM}=\frac{\sum_{i\in \mathcal{B}}(y_{i}-\hat{y}^{*}_{i})^{2}-\sum_{i\in \mathcal{B}}(y_{i}-\hat{y}_{i})^{2}}{|\mathcal{B}|},
\end{equation}
where $\mathcal{B}$ is the test samples used to evaluate the importance scores and select variables. $y_{i}$ is the ground truth label,  $\hat{y}_{i}$ and
$\hat{y}^{*}_{i}$ are the predictions of any given model for data $i$ 
before and after permuting the selected variable, respectively. 
The main idea of PVIM is to measure how much the prediction  accuracy will
decrease without a variable’s modeled information.
Therefore, if a model can adequately model a variable's main effect
and all involved interactions, PVIM could perfectly measure its overall
importance. However, the potential of PVIM remains underutilized, as the state-of-the-art literature often pays limited attention to thoroughly modeling all interaction effects associated with a single important variable. To address this gap, we propose designating an important variable as the \emph{King} and using \emph{King’s PVIM} to identify and gather all variables that share at least one interaction with the \emph{King}, collectively referred to as the \emph{King’s Core Team}, and subsequently measuring the \emph{King’s} overall importance.


In the current state-of-the-art literature, one of the most inspiring approaches to addressing this gap is iteratively reweighting variable importance using the mean decrease in impurity\citep{archer2008empirical,basu2018iterative}. By incrementally adding the impurity decrease of the selected variable to its weight at each node, the iterative Random Forests (iRF) algorithm progressively prioritizes important variables with higher weights for node splitting. This adjustment increases the likelihood of selecting these variables, moving them closer to the root node, and grouping interaction participants within the same tree path. Moreover, the reweighting process naturally facilitates variable selection by ranking variables based on their updated importance. Although iteratively reweighted techniques show great potential for exploring interactions, the following three limitations present opportunities for improvement. These challenges motivate us to propose a novel method that efficiently uncovers interactions and analyzes their structure, as detailed in the next subsection:

\begin{enumerate}[left=0pt]
    \item \emph{Lack of Incorporation of Prior Knowledge}: Existing methods do not fully leverage prior knowledge to enhance interaction discovery. When a variable is known to be important, placing it at the root node of trees can facilitate the identification of other variables that interact with it. This concept is central to our approach, where a \emph{King} is selected to identify its \emph{Core Team}.
    \item \emph{Suboptimal Criterion for Weight Updating}: Impurity decrease, as a training-phase metric, is not the most effective criterion for updating variable weights. During training, a random subset of variables is tested to select the one with the largest impurity decrease for node splitting. In cases with a large number of variables ($p$) and sparse truly important ones, important variables may not be selected during many node splits, while uninformative variables might coincidentally exhibit large impurity decreases, ultimately accumulating high weights. To overcome this limitation, we introduce a testing-phase metric, \emph{King's PVIM}, which better identifies informative trees and prioritizes important variables.
    \item \emph{Absence of Interaction Structure Evaluation}: Current techniques lack criteria to analyze the structure of interactions. This motivates us to propose two inference metrics designed to detect interactions and evaluate their order and types.
\end{enumerate}

\subsection{Iterative Kings' Forests}

\begin{figure}[h!]
   \centering    \includegraphics[width=0.9\textwidth]{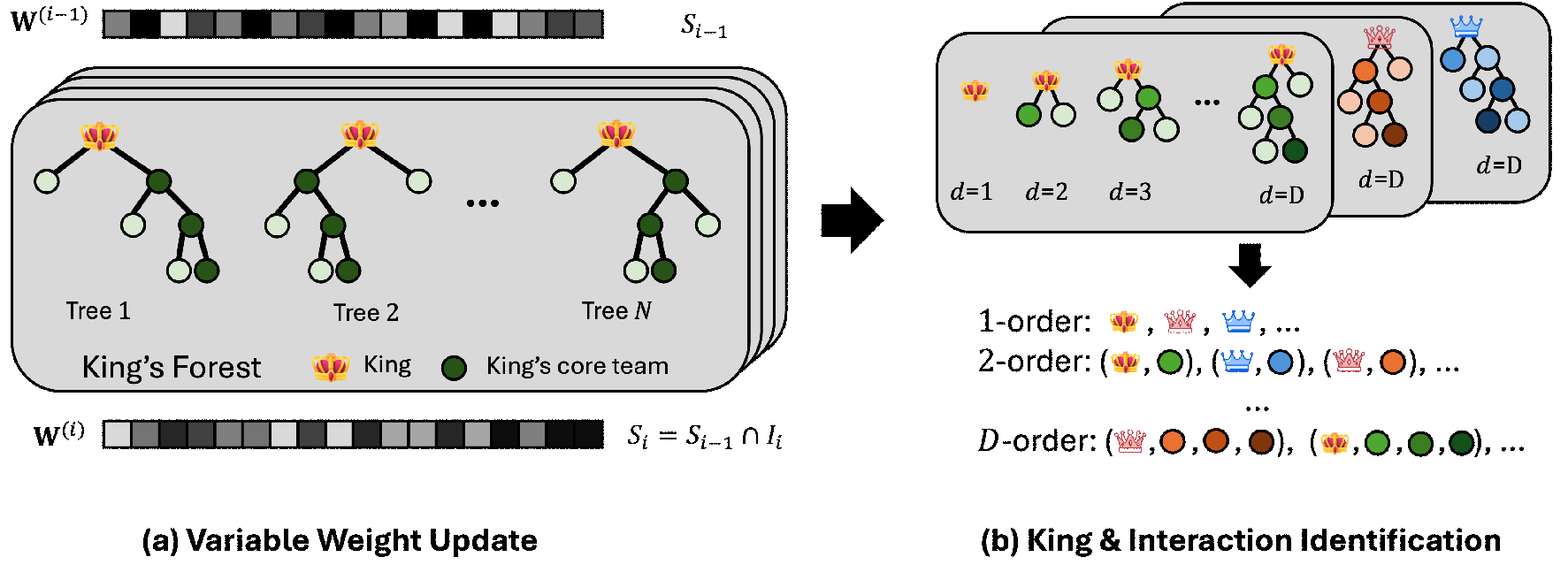}
    \caption{iKF Overview: Iteratively select new Kings until $|S_{i}|$ is sufficiently small. See definition of $S_{i}$ and $I_{i}$ in step (4) below. (a) Use a series of disposable King's Forests to update the variables' weights $\vw^{(i)}$, and rank $\vw^{(i)}$ to select the Core Team for the $i$-th King. (b) Build $D$ King's Forests to uncover $d$-order interactions by analyzing depth-$d$ tree paths for the $i$-th King.}
    \label{fig:ikf_overview}
\end{figure}

In this section, we propose a novel method, Iterative Kings' Forests (iKF), designed to uncover interactions among variables. 
An overview of iKF is illustrated in Figure~\ref{fig:ikf_overview}.

To start, prior knowledge can guide the selection of an important variable as the first \emph{King}. In the absence of prior knowledge, a random variable can be selected instead. After choosing the first \emph{King}, its corresponding \emph{King's Forests} are constructed by fixing it as the root node for all trees in the forest. The key concepts of \emph{King's Forests} are detailed in steps 1 through 3. Following this, iKF iteratively determines the most suitable variable to act as the next \emph{King} and constructs new \emph{King's Forests} for it. This iterative process continues until a predefined stopping criterion is reached. Key aspects of this procedure are elaborated on in step 4.




\textbf{(1) Tree selection \& Variable weight update.} We compute King's PVIMs for every tree in a \emph{King's Forest} and only select trees with positive King's PVIMs as informative trees. A larger King's PVIM indicates a tree does better in selecting the \emph{King’s Core Team}, and modeling the interactions about the King. Therefore, we use King's PVIM to quantify the importance of a tree, then add the King's PVIM to the weight of all variables selected in this tree. An informative tree could include the members of \emph{King’s Core Team}, other important variables, and some random noise variables. 

We initialize an equal weight for all the variables as $\vw^{(0)}=\textbf{1}\in\mathbb{R}^p$, and then iteratively update variables’ weight for each variable $x_{i}$ within all the trees with positive King's PVIM:
\begin{equation}
w_{i}^{(t)} = w_{i}^{(t-1)}+\sum_{j=1}^{N}\text{PVIM}_{j}*I(\text{PVIM}_{j}>0)*I(x_{i}\text{ is in j-th tree}),
\label{eq:weight_update}
\end{equation}
where N is forest size and iteration $t=1,...,N_{\text{iter}}$. In this way, the members of \emph{King’s Core Team} could gradually gain larger weights, and therefore more likely to move toward the root node.



\textbf{(2) Variable selection \& Depth-$d$ tree paths.} 
After updating the variable weights, we select the top $N_{c}$ variables as candidates for the \emph{King’s Core Team} and construct $D$ final \emph{King’s Forest}s with maximum depths $d=1,2,...,D$, aiming to recover the King’s $d$-order interactions from all depth-$d$ paths. For a forest of size $N$ and depth $d$, it contains $l$ $(l\leq N\times2^{d-1})$ distinct paths, or indicator functions, representing potential $d$-order interactions. 

As noted earlier, an informative tree could include the members of \emph{King’s Core Team}, other important variables, and some random noise variables. Consequently, a depth-$d$ path does not necessarily represent a true $d$-order interaction. 

\textbf{(3) Evaluation metrics \& Interaction inference.} 
To better identify interactions, we introduce the following two inference metrics to evaluate a given depth-$d$ path:
\begin{enumerate}[left=0pt]
    \item \emph{King's PVIM}: The sum of \emph{King}'s PVIMs across all trees containing the given path.
    \item \emph{Path Reproduction Count}: The number of times the path is reproduced in the forest.
\end{enumerate}
\textbf{Notes:} Dividing \emph{King's PVIM} by the \emph{Path Reproduction Count} produce the average \emph{King's PVIM} of a path (potential interaction), offering alternative refined metrics to assess potential interactions.


For each depth $d$, the King’s Forests algorithm generates two depth-$d$ path lists, each of size $N_{\text{top}}$, ranked according to the two inference metrics. A path is considered a promising candidate of an active order-$d$ interaction if it ranks highly in at least one of the two lists. 

Among the two metrics, the testing criterion, \emph{King's PVIM}, is effective for evaluating the order of a \emph{King}’s interaction by analyzing the trend of \emph{King}’s PVIMs as the maximum depth $d$ increases. For instance, if a \emph{King}’s PVIM shows a substantial increase from $d=2$ to $d=3$, it suggests that the \emph{King} likely participates in at least one order-$3$ interaction not captured by forests with maximum depth 1 or 2. Conversely, if no such increase is observed, the \emph{King} does not contribute to a third-order interaction. A large \emph{Path Reproduction Count} indicates that all variables in the path are likely important, but it does not necessarily imply they form a single interaction; they could originate from two separate interactions or a combination of a main effect and an interaction. 

We will illustrate the process of uncovering \emph{King's} interactions using these two metrics in detail with examples in Sections 4.3 and 5.3. \textbf{King’s Forests} Algorithm is outlined in Algorithm \ref{alg:1}.

\begin{algorithm}[ht!]
\caption{: King's Forests}
\label{alg:1}
\begin{algorithmic}[1]
\STATE {\bfseries Input:} Forest size $N$, maximum depth $D$, \emph{King}
\STATE Initialize $\vw^{(0)} = \mathbf{1}$

\# Update variable weights
\FOR{$t=1$ {\bfseries to} $N_{\text{iter}}$}

\STATE Construct a \emph{King}'s forest of size $N$ and depth $D$
    \FOR{tree $j$ in the \emph{King}'s forest}
        \FOR{each splitting variable $x_{i}$ in tree j}
        \STATE Increase the weight of $x_{i}$ by \emph{King}’s PVIMs of tree j as in Eq.~\ref{eq:weight_update}:
        $$w_{i}^{(t)} = w_{i}^{(t-1)}+ \text{PVIM}_{j}*I(\text{PVIM}_{j}>0)*I(x_{i}\text{ is in j-th tree})$$
        \ENDFOR
    \ENDFOR
\ENDFOR

\# Interaction identification.
\STATE Select the top $N_{c}$ variables to construct $D$ King’s forests with maximum depth $d$ from  1 to $D$.
\STATE Calculate two inference metrics of all paths from depth 2 to $D$.
\STATE For each depth $d$ (2 to $D$), identify the top $N_{\text{top}}$ depth-$d$ paths ranked by each inference metric.
\STATE {\bfseries Output:} Variable weights $\vw^{(N_{\text{iter}})}$, \emph{King's PVIMs} for orders 1 to $D$, and depth-$d$ path lists for two inference metrics with $d$ from 2 to $D$.
\end{algorithmic}
\end{algorithm}

\textbf{(4) Iterative Kings' Forests.} The \emph{King’s Forests} algorithm focuses exclusively on the interactions involving a single \emph{King}. To achieve a comprehensive view, we iteratively select the next \emph{King}, the variable with the highest accumulated weight $\vW$ among those not yet designated as a \emph{King}, and construct its corresponding \emph{King’s Forests}. This iterative process, known as \emph{Iterative King's Forest} (iKF), continues until a predefined stopping criterion is met. Notably, the stopping criterion is not fixed and serves primarily to determine when to conclude the automatic process of selecting the next King and constructing new King's Forests. Users can tailor this criterion to their needs, such as specifying the desired number of Kings to select in iKF and using that number as the stopping condition. The detailed steps of the \textbf{Iterative Kings' Forests (iKF)} algorithm are outlined in Algorithm \ref{alg:2}.

\textbf{Stopping criterion.} Starting with all $p$ variables, denoted as $S_{0}$, iKF ranks the weights $\mathbf{w}^{(i)}$ in descending order during the $i$-th iteration and removes the bottom $\alpha$ (e.g., $20\%$) of variables. The remaining top $1 - \alpha$ (e.g., $80\%$) variables form the set $I_{i}$, representing those that survived the $i$-th iteration. The survived variable set after the $i$-th iteration is then calculated as $S_{i} = S_{i-1}\cap I_{i}$, representing the important variables that persist through the first $i$ iterations. This process continues, iteratively selecting a new King and constructing King’s Forests, until the size of $S_{i}$ falls below a pre-specified threshold $K$. 



\begin{algorithm}[ht]
\caption{: Iterative Kings' Forests (iKF)}
\label{alg:2}
\begin{algorithmic}[1]
\STATE {\bfseries Input:} Forest size $N$, maximum depth $D$, first \emph{King} $x^{(1)}$
\STATE Initialize weights $\mathbf{W}^{(0)} = \mathbf{0}$, and the survived variable set $S_0 = \{1,\ldots,p\}$. 

\WHILE{$|S_{i}|> K $}
\STATE For $i$-th \emph{King} $x^{(i)}$, implement Algorithm \ref{alg:1} to obtain weight $\vw^{(i)}$, \emph{King's PVIMs}, and all path lists.
\STATE Update survived variable set: $S_i = S_{i-1}\cap I_{i}$.
\STATE Update weight as $\vW^{(i)} = \vW^{(i-1)} + \vw^{(i)}$.
\STATE Among those not previously designated as a \emph{King}, select the variable with the highest weight in $\vW^{(i)}$ as the new \emph{King} $x^{(i)}$ 
\STATE $i\leftarrow i+1$
\ENDWHILE

\STATE For all selected Kings, concatenate their \emph{King's PVIMs} for orders 1 to $D$, and concatenate their depth-$d$ path lists for each metric and maximum depth $d$ (2 to $D$).
\STATE {\bfseries Output:} Variable final weights $\vW$, concatenated \emph{King's PVIMs} for orders 1 to $D$, and concatenated depth-$d$ path lists for each metric with $d$ from 2 to $D$.

\end{algorithmic}
\end{algorithm}

\subsection{Scientific Interaction Discovery}

While it is unrealistic for iKF to identify all interactions and their structures with 100\% accuracy, we propose the following criteria to evaluate iKF outcomes. These criteria aim to illuminate hidden interaction mechanisms, thereby enhancing interpretability.

\begin{itemize}[left=0pt] 
    \item \textbf{Overall Variable Importance.} The final weights $\vW$ are utilized for variable ranking and selection, emphasizing the relative importance of each variable.  
    \item \textbf{Kings' PVIMs.} For each King, the progression of \emph{King's PVIMs} from order 1 to $D$ reflects the interaction orders involving the King. For example, if a \emph{King's PVIM} increases from order 1 to 2 and then from order 2 to 3, this pattern indicates the King's involvement in at least one second-order interaction and one third-order interaction.  
    \item \textbf{Interaction Candidate Shortlists.} A path is considered a strong candidate for an active order-$d$ interaction if it ranks highly in at least one of the two depth-$d$ path shortlists generated using inference metrics.
    \item \textbf{Path Directions.} If both paths $(x_1, x_2)$ and $(x_2, x_1)$ appear in the shortlists, it indicates that $x_1$ and $x_2$ have equal status. If only one direction is selected, the interaction is likely hierarchical. Here, $x_1$ and $x_2$ may represent either a single variable or an interaction.
\end{itemize}

\textbf{Three Interaction Types.} We categorize interactions into three types to enhance model interpretability. The first type includes interactions that satisfy the weak heredity condition, where at least one variable shows an accompanied main effect. The second type includes interactions where all variables involved have equal status. Finally, the third type consists of interactions with a nested structure. Their definitions are as follows:

\begin{enumerate}[left=0pt] 
\item \textbf{Accompanied Interactions.} At least one variable in the interaction exhibits a main effect.
\item \textbf{Synergistic Interactions.} The interaction lacks an active main effect, with all variables contributing collectively, but not independently. 
\item \textbf{Hierarchical Interactions.} Interactions follow a specific sequence, where a dominant variable must be modeled first to activate a nested variable, rather than the reverse. This ordering is especially crucial in biological research, where, for instance, genes typically function in a sequence, with upstream genes activating the expression of downstream genes.
\end{enumerate}

iKF most readily identifies Accompanied Interactions due to their associated main effects. Simultaneously, it gradually assigns higher weights to participants in Synergistic and Hierarchical Interactions, enhancing the likelihood of selecting these participants as "Kings" and detecting the interactions they are part of after several iterations. Furthermore, iKF differentiates between Synergistic and Hierarchical Interactions by analyzing the Kings' PVIMs and the order of variables in the selected interactions.

%% file: section/exp.tex
\section{Interaction Discovery in Experimental Scenarios}

To evaluate the performance of our iKF in variable selection and interaction identification, we conduct simulation experiments using various ground-truth functions to assess whether iKF accurately recovers multi-order interaction effects.

\subsection{Experimental Settings}

\textbf{Simulated functions.} We simulate two categories of functions:  In category \textbf{(a)}, four variables influence the response through two pairwise interactions of different forms, while in category \textbf{(b)}, three variables affect the response through one third-order interaction. Their ground-truth functions, assuming both the error term $e$ and any variable $x_{i}$ follow a standard normal distribution, are defined as follows:
\begin{description}
  \item \textbf{(a1):} $y = 2*x_{1}*x_{3} - 2*x_{5}*(x_{7}<0.2) + e$; \textbf{(b1):} $y = 2*x_{1}*(1+x_{3})^{2}*\sin(x_{5}) + e$
  \item \textbf{(a2):} $y = 4*x_{1}*\sin(x_{3}) - 4*x_{5}*\cos(x_{7}) + e$; \textbf{(b2):} $y=2*x_{1}*sign(1+x_{3})*\sin(x_{5})+e$ 
  \item \textbf{(a3):} $y = 2*x_{1}*x_{3} + 3*x_{5}*\sin(x_{7}) + e$; \quad\; \textbf{(b3):} $y = 2*x_{1}*x_{3}*x_{5} + e$
\end{description}

\textbf{Baselines.} We compare the performance of our iKF with a classical variable selection method: DC-SIS~\citep{li2012feature}, and a standard interaction identification method: iRF~\citep{basu2018iterative}.

\textbf{Hyperparameters setting.}
We set the sample size $n = 200$ and the number of all variables $p = 500$. 
We conduct 100 replications to update the variable weight list and identify the interactions.
For iKF, we set $N_{c} = \lfloor n/2\log(n)\rfloor$, $\alpha = 0.5$, $N_{\text{iter}} = 7$, the forest size $N = 100$, maximum depth $D = 4$ for part \textbf{(a)} and $D = 5$ for part \textbf{(b)}, and the size of depth-d path shortlists $N_{\text{top}} = 20$.

\textbf{Evaluation metrics.}
We evaluate the performance of variable using the following criteria: (1)
  Minimum recovery size (\textbf{MRS}): It refers to the minimum size of the output ranked variable list required to include all the important variables. Therefore, a smaller $\mathcal{S}$ indicates greater recovery power of the model.
  (2) Individual recovery rate (\textbf{IRR}): The proportion that one specific interaction effect is selected in 100 replications.
  (3) Overall recovery rate (\textbf{ORR}): The proportion that all active interaction effects are selected in 100 replications.

\subsection{Main Experiment}

In this section, we will include two parts to evaluate the  performance of \emph{variable selection} and \emph{interaction identification}.
Specifically, we will assess the iKF’s performance in variable selection through comparison with DC-SIS and iRF using minimum recovery size (MRS), and compare iKF and iRF in identifying interaction effects through individual recovery rate (IRR \%) for each active interaction and overall recovery rate (ORR \%) for all the active interactions.

\begin{figure}[h!]
  \centering

    \includegraphics[width=0.48\textwidth]{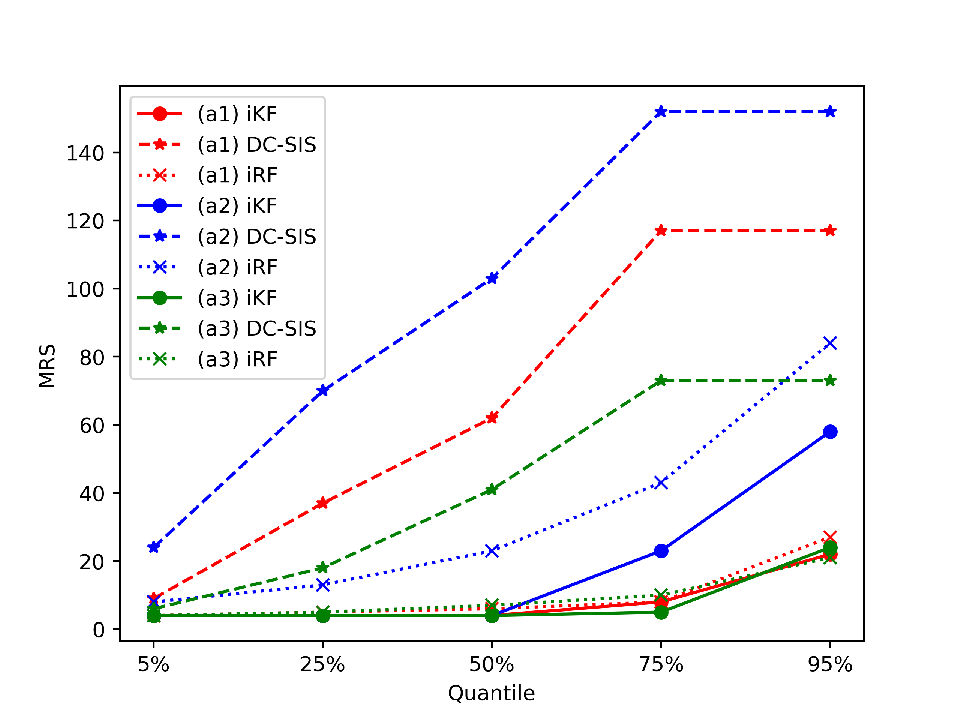}
     \includegraphics[width=0.48\textwidth]{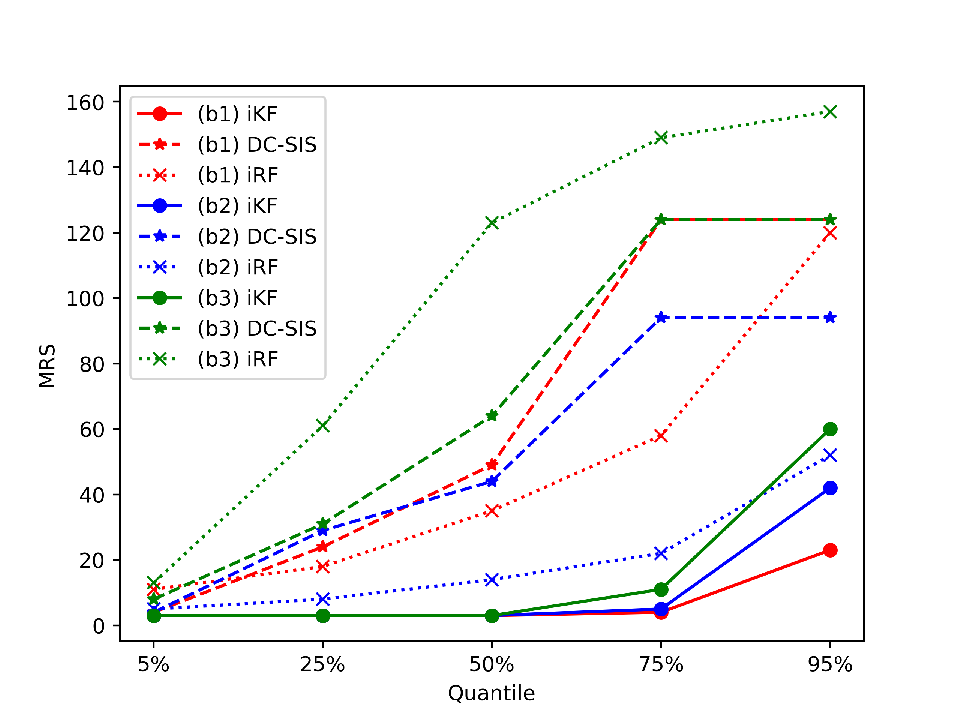}
  \caption{Minimum recovery size (MRS) across various quantiles.}\label{fig1}
  \vspace{-3mm}
\end{figure}

\textbf{Variable selection.} 
For each replicated experiment, we obtain a variable weight list and determine an MRS that covers all active variables. After conducting 100 replications, we collect 100 MRS values, rank them, and report the 5\%, 25\%, 50\%, 75\%, and 95\% quantile values. From the results in Figure~\ref{fig1}, we can observe that our iKF can outperform DC-SIS and iRF across various quantiles in all simulated functions.

\textbf{Interaction identification.} Besides variable selection, we are more interested in the capability to recover interactions. We  quantify this capbility with 
IRR (\%) for specific interaction and ORR (\%) for all the related interactions. From the results in Table~\ref{t2}, we can make the following observations:
\begin{itemize}[left=0pt]
    \item In Part (a), iKF can significantly outperforms iRF in identifying both the individual and overall interactions.
    \item In Part (b), iRF fails to identify all three third-order interactions, while iKF has a more than 48\% probability to discover them in cases (b1) - (b2).
    \item Our iKF has greater power to identify important interactions, particularly those of higher orders.
\end{itemize}

\begin{table}[ht]
    \centering
    \begin{subtable}{0.45\textwidth} 
        \centering
\begin{tabular}{ccccccccccccc}
\toprule
\textbf{Task} & \textbf{Method} & $\textbf{IRR}_{13}$ $\uparrow$ & $\textbf{IRR}_{57}$ $\uparrow$ & \textbf{ORR} $\uparrow$  \\
\midrule
\multirow{2}{*}{\textbf{(a1)}} & iRF &  0.12 & 0.84 & 0.09   \\
&iKF &0.73 & 0.82 & 0.60\\\hline
\multirow{2}{*}{\textbf{(a2)}} &iRF & 0.19 & 0.15 & 0.02   \\
&iKF&0.53 & 0.73 & 0.41\\\hline
\multirow{2}{*}{\textbf{(a3)}} &iRF &  0.28 & 0.14 & 0.04  \\
&iKF&0.57 & 0.45 & 0.28 \\
\bottomrule
\end{tabular}
        \caption{Result in Part (a).}
    \end{subtable}
    \hspace{1.4cm} 
    \begin{subtable}{0.4\textwidth}
        \centering
\begin{tabular}{ccccccccccccc}
\toprule
\textbf{Task}& \textbf{Method}  & \textbf{ORR} $\uparrow$\\
\midrule
\multirow{2}{*}{\textbf{(b1)}} & iRF &  0.0  \\
&iKF &0.68 \\\hline
\multirow{2}{*}{\textbf{(b2)}} &iRF & 0.0  \\
&iKF&0.48 \\\hline
\multirow{2}{*}{\textbf{(b3)}} &iRF &  0.0  \\
&iKF&0.05  \\
\bottomrule
\end{tabular}
        \caption{Result in Part (b).}
    \end{subtable}
    \caption{Interaction identification for both individual (\textbf{IRR}) and overall (\textbf{ORR}) interactions. Our iKF can significantly outperform iRF across all the settings. (\textbf{IRR}$_{13}$ and \textbf{IRR}$_{57}$ represent the individual recovery rate for $(x_1,x_3)$-interaction and $(x_5,x_7)$-interaction, respectively.) }
    \label{t2}
\end{table}



\subsection{Case Study}
In the previous section, we evaluated the overall performance of different methods in variable selection and interaction identification. However, it is still unclear how our iKF works. To address this, we use two examples in Parts (a) and (b) to delve into the working mechanism of iKF and demonstrate the procedure for selecting important variables, identifying interactions, and categorizing interactions.

\textbf{Case 1 from Part (a1):} $y=2*x_{1}*x_{3}-2*x_{5}*(x_{7}<0.2)+e$

We apply iKF in case 1 to output \emph{King}s' PVIMs
and
top ten tree paths ranked by sum of PVIM for exploring the hidden model mechanism.
\begin{table}[ht!]
\caption{\label{a1} Kings' PVIMs across all the depths for case 1.}\vspace{0.2cm}
\centering
\scalebox{0.9}{
\begin{tabular}{ccccccccc}
\toprule
 King $\backslash$ Depth  & $d=1$ & $d=2$ & $d=3$ & $d=4$\\ 
  \hline
 $x_5$   & 4.65  & 5.67 & 5.72 & 6.34 \\ 
 $x_7$   & -0.21 & 1.79 & 1.26 & 0.93 \\ 
 $x_3$  & 0.19  & 3.75 & 4.03 & 4.47 \\ 
 $x_1$   & -0.16 & 3.48 & 5.23 & 4.44  \\ 
 $x_{10}$  & -0.19 & 0.33 & 0.33 & 0.49 \\ 
 $x_{146}$ & 0.18  & 0.59 & 0.53 & 0.34 \\ 
 $x_{91}$  & 0.21  & 0.31 & 0.77 & 0.16 \\ 
   \bottomrule
\end{tabular}
}
\end{table}
The iKF chooses seven variables as \emph{King}s and provides corresponding  PVIMs from depth $d$ from 1 to 4 in Table \ref{a1}. When the depth is 1, Kings' PVIMs show that variables 3, 5, 91, and 146 may have main effects, among which the variable 5 has a large main effect. When the depth increases to 2, PVIMs of variables 1, 3, 5, and 7 sharply increase. This indicates that all four variables participate in some pairwise interactions.

Our iKF also ranks variables based on variables' final weights, which give the top 15 variables as $(x_5, x_7, x_3, x_1, x_{110}, x_{104}, x_{193}, x_{48}, x_{146}, x_{122}, x_{49}, x_{103},  x_{47}, x_{190},  x_{64}$). This result shows that the four important variables, i.e., $x_1,x_3,x_5,x_7$,  are ranked as the top four variables.

\begin{table}[ht]
\caption{\label{a2} Top ten depth-2 tree paths for case 1.}\vspace{0.2cm}
\centering
\scalebox{0.9}{
\begin{tabular}{cccccccccc}
\hline
Rank & Depth 1 & Depth 2 & \# of Recovery & Sum of PVIMs \\ 
\hline
1  & $x_5$ & $x_7$ & 72 & 493 \\ 
  2 & $x_1$ & $x_3$ & 93 & 434 \\ 
  3  & $x_3$ & $x_1$ & 88 & 419 \\ 
  4 & $x_1$ & $x_5$ & 72 & 286 \\ 
  5  & $x_3$ & $x_5$ & 82 & 280 \\ 
  6  & $x_7$ & $x_5$ & 91 & 202 \\ 
  7  & $x_5$ & $x_{104}$ & 15 & 96 \\ 
  8  & $x_5$ & $x_3$ & 18 & 85 \\ 
  9  & $x_{146}$ & $x_5$ & 98 & 84 \\ 
  10 & $x_{91}$ & $x_5$ & 76 & 81 \\ 
 \hline
\end{tabular}
}
\end{table}

Table \ref{a2} gives the top ten depth-2 tree paths as possible pairwise interactions. Among them, ($x_5$, $x_7$) is selected in both directions, path ($x_5$, $x_7$) and path ($x_7$, $x_5$). That is, when the \emph{King} is $x_5$, it tends to select variable $x_7$ to split the node of the next layer, and vice versa. Similar things happen to ($x_1$, $x_3$). 

Based on the results from Tables~\ref{a1} and \ref{a2}, we conclude that the variables $x_1$, $x_3$, $x_5$, and $x_7$ are important, with $x_5$ showing a main effect on the response. Additionally, iKF identifies ($x_1$, $x_3$) and ($x_5$, $x_7$) as important pairwise interactions, where ($x_1$, $x_3$) is classified as a Synergistic Interaction and ($x_5$, $x_7$) as an Accompanied Interaction.

\textbf{Case 2 from Part (b2):} $y=2*x_{1}*sign(1+x_{3})*\sin(x_{5})+e$\\
Similarly, we also apply iKF on case 2 to output Kings' PVIMs and top ten tree paths.  
\begin{table}[ht!]
\caption{\label{b1} Kings' PVIMs across all the depths for case 2.}\vspace{0.2cm}
\centering
\scalebox{0.9}{
\begin{tabular}{ccccccccc}
\toprule
 King $\backslash$ Depth  & $d=1$ & $d=2$ & $d=3$ & $d=4$\\ 
  \hline
 $x_{63}$ & -0.03 & 0.15 & -0.07 & -0.03 \\ 
 $x_{1}$  & 0.20 & 1.11 & 1.33 & 1.24 \\ 
 $x_{5}$  & -0.10 & 1.12 & 1.23 & 1.17 \\ 
 $x_{3}$  & 0.06 & 0.06 & 0.12 & 0.25 \\ 
 $x_{46}$ & 0.12 & 0.14 & 0.16 & 0.10 \\ 
   \hline
\end{tabular}
}
\end{table}
The iKF chooses five variables as ``Kings'' and give corresponding Kings' PVIMs from depth 1 to 4 in Table \ref{b1}. When the depth is 1, Kings' PVIMs show that no variable has a strong main effect. PVIMs of variables 1 and 5 sharply increase when the depth increases to 2, and keep increasing when the depth reaches 3. This indicates that 3 and 5 are involved in pairwise interactions and possibly third-order interactions.

The iKF also ranks variables based on variables' final weights. The top 15 variables, $(x_{1}$ $, x_{5}$ $, x_{3}$ $, x_{69}$ $, x_{114}$ $, x_{46}$ $, x_{98}$ $, x_{24}$$, x_{10}$ $, x_{6}$ $, x_{189}$ $, x_{68}$$, x_{131}$ $, x_{15}$ $, x_{182})
$, show that the three important variables, $x_{1}, x_{5}, x_{3}$, are ranked as the top three.

\begin{table}[ht]
\caption{\label{b2} Top ten depth-2 tree paths for case 2.}\vspace{0.2cm}
\centering
\scalebox{0.9}{
\begin{tabular}{cccccccccc}
\toprule
Rank & Depth 1 & Depth 2 & \# of Recovery & Sum of PVIMs \\ 
  \hline
1 & $x_{1}$ & $x_{5}$ & 123 & 189 \\ 
  2  & $x_{5}$ & $x_{1}$ & 126 & 161 \\ 
  3  & $x_{5}$ & $x_{27}$ & 12 & 14 \\ 
  4  & $x_{1}$ & $x_{69}$ & 11 & 10 \\ 
  5  & $x_{1}$ & $x_{121}$ & 5 & 9 \\ 
  6  & $x_{3}$ & $x_{69}$ & 39 & 9 \\ 
  7 & $x_{5}$ & $x_{194}$ & 5 & 8 \\ 
  8  & $x_{1}$ & $x_{83}$ & 5 & 8 \\ 
  9 & $x_{5}$ & $x_{190}$ & 5 & 8 \\ 
  10  & $x_{5}$ & $x_{28}$ & 5 & 7 \\ 
\bottomrule
\end{tabular}
}
\end{table}

In Table \ref{b2}, ($x_{1}$, $x_{5}$) is ranked among the top two in both directions, path ($x_{1}$, $x_{5}$) and path ($x_{5}$, $x_{1}$), indicating that ($x_{1}$, $x_{5}$) is an important interaction. It is likely a Synergistic Interaction, as neither $x_{1}$ or $x_{5}$ shows a main effect or nested effect.

\begin{table}[ht]
\caption{\label{b3} Top ten depth-3 tree paths for case 2.}\vspace{0.2cm}
\centering
\scalebox{0.9}{
\begin{tabular}{cccccccccc}
\toprule
Rank & Depth 1 & Depth 2 &Depth 3 & \# of Recovery & Sum of PVIMs \\  
  \hline
 1 & $x_{5}$ & $x_{1}$ & $x_{3}$ & 39 & 66 \\ 
  2  & $x_{1}$ & $x_{5}$ & $x_{3}$ & 17 & 40 \\ 
  3 & $x_{1}$ & $x_{5}$ & $x_{184}$ & 12 & 24 \\ 
  4  & $x_{1}$ & $x_{5}$ & $x_{69}$ & 18 & 23 \\ 
  5  & $x_{1}$ & $x_{5}$ & $x_{200}$ & 14 & 20 \\ 
  6  & $x_{1}$ & $x_{5}$ & $x_{24}$ & 9 & 18 \\ 
  7 & $x_{5}$ & $x_{128}$ & $x_{1}$ & 18 & 18 \\ 
  8  & $x_{5}$ & $x_{1}$ & $x_{69}$ & 10 & 16 \\ 
  9  & $x_{1}$ & $x_{5}$ & $x_{7}$ & 6 & 16 \\ 
  10  & $x_{1}$ & $x_{121}$ & $x_{5}$ & 6 & 15 \\ 
\bottomrule
\end{tabular}
}
\end{table}

Table \ref{b3} lists the top ten depth-$3$ tree paths. Notably, ($x_{1}$, $x_{3}$, $x_{5}$) ranks among the top two in two directions, ($x_{1}$, $x_{5}$, $x_{3}$) and ($x_{5}$, $x_{1}$, $x_{3}$). This strongly suggests that ($x_{1}$, $x_{3}$, $x_{5}$) is an important third-order interaction.

Based on results from Tables \ref{b1}-\ref{b3}, iKF concludes that variables $x_{1}$, $x_{3}$ and $x_{5}$ are important variables, and no variable has a main effect on the response. Moreover, iKF indicates that ($x_{1}$, $x_{5}$) is an important pairwise interaction, and ($x_{1}$, $x_{3}$, $x_{5}$) is an important third-order interaction. For interaction types, ($x_{1}$, $x_{5}$) is categorized as Synergistic Interaction. When treating the interaction ($x_{1}$, $x_{5}$) as a single variable, iKF classifies the interaction ($x_{1}$, $x_{5}$, $x_{3}$) as a Hierarchical Interaction, with ($x_{1}$, $x_{5}$) as a dominant variable and $x_{3}$ as a nested variable. This is because the King’s PVIM for $x_{3}$ increases only slightly with depth, and depth-3 paths ($x_{3}$, $x_{1}$, $x_{5}$) and ($x_{3}$, $x_{5}$, $x_{1}$) are not selected in Table \ref{b3}.


\section{Biological Scientific Discovery}

In the previous section, we examined iKF's ability to identify important variables and interactions across various experimental scenarios. In this section, we further demonstrate how iKF operates and highlight its potential for scientific discovery in biological research by applying it to the \emph{Drosophila embryo} case study.

\subsection{Dataset and Hyper-parameter Configuration}

\textbf{Dataset.} 
\emph{Drosophila embryo} dataset~\citep{basu2018iterative} consists of 7809 genomic sequences with enhancer status as a binary response, where the sequences driving patterned expression in blastoderm (stage 5) embryos has a positive label. In the early Drosophila embryo, patterning is driven by interactions among about 40 TFs \citep{rivera1996gradients}, which makes it a valuable test case to evaluate the performance of iKF.

\textbf{Hyperparameters.}
We set $N_{c} = \lfloor p/2 \rfloor$, $\alpha = 0.2$, $N_{\text{iter}} = 6$, the maximum depth $D = 5$, forest size $N = 200$ and the size of depth-d path shortlists $N_{\text{top}} = 30$. 

\subsection{Gene Expression Regulation in Drosophila Embryogenesis}
Precisely regulated Spatio-temporal gene expression is crucial for multi-cellular organism development, where enhancers play a critical role by coordinating combinatorial transcription factor (TF) binding. These activities lead to patterned gene expression during embryogenesis \citep{levine2010transcriptional}. One of the best-studied developmental embryogenesis cases is Drosophila embryo in which TF hierarchies act to pattern and subdivide the embryo along the anteroposterior (AP) and dorsoventral (DV) body axes.
The zinc-finger protein \emph{Zelda (Zld)} plays a key role as an early regulatory factor in timing zygotic gene activation and promoting robust expression \citep{liang2008zinc}. When lacking maternal expression of \emph{Zld} in early embryos, expression profiling studies reveal that many genes normally activated between 1-2 hrs of development are strongly down-regulated and never recovered, including genes related to cellularization, sex determination, and dorsal patterning \citep{liang2008zinc}. 
\citet{nien2011temporal} points out that \emph{Zld} binds to 72\% of the \emph{Bcd} targets, 70\% of the \emph{Cad} targets, and 80\% of the \emph{Tll} targets. 
About 50\% overlap is observed between \emph{Zld} targets and gap gene (\emph{Hb, \emph{Gt}, \emph{Kr}, \emph{Kni}}) targets. 
\emph{Zld} also regulates the expression by orchestrating the timing within the segmentation gene network \citep{nien2011temporal}. In the absence of \emph{Zld}, embryos initial transcription of gap-genes is delayed by 1-2 nuclear cycle (nc). In addition, their patterns are significantly disrupted, which can be explained by miscued gap-gene interactions.
Meanwhile, without \emph{Zld}, some genes involved in ventral patterning, for example \emph{\emph{Twi}}, is just temporally delayed, but later recovered by nc 14 \citep{nien2011temporal}. \emph{\emph{Twi}} functions as a basic helix-loophelix (bHLH) activator, and is essential for specifying the ventral neurogenic ectoderm. At least half of the tissue-specific enhancers regulated by Dorsal also contain binding sites for \emph{\emph{Twi}} \citep{stathopoulos2002whole,markstein2004regulatory}.

\subsection{Rediscovery of TF Interaction Mechanisms}

To validate the gene expression regulation mechanism as shown in previous section, we conduct the iKF procedure for interaction identification using early \emph{Drosophila embryo} data. In this section, we will present identified interaction lists and Kings identification mechanism in iKF.

\textbf{TF interactions identification Comparison with iRF.} 
iRF \citep{basu2018iterative} outputs its top $20$ pairwise TF interactions, in which $16$ interactions are verified in biological literatures\citep{harrison2011zelda,nien2011temporal,li2008transcription,kraut1991spatial,eldon1991interactions,struhl1992control,capovilla1992giant,schulz1994autonomous,zeitlinger2007whole,nguyen1998drosophila,hoch1991gene,hoch1990cis}. For comparison, our iKF can identify $14$ pairwise interactions out of the $16$ verified interactions, and 2 pairwise interactions out of the 4 unverified interactions. This significant overlap verifies the capacity of iKF in identifying influential interactions. 

Moreover, iKF identifies additional interactions overlooked by iRF, including three pairwise interactions and one third-order interaction, all of which we have verified in biological literature.

\begin{itemize}[left=0pt] 
    \item \textbf{Pairwise Interaction.} \citet{harrison2011zelda} and \citet{nien2011temporal} showed that \emph{Kni}, \emph{Ftz} and \emph{Cad} are TFs regulated by \emph{Zld}, forming three pairwise interactions involving \emph{Zld}.
    \item \textbf{Third-order Interaction.} (\emph{Zld}, \emph{Kni}, \emph{Tll}), is a third-order interaction. As \citet{moran2006tailless} noted, \emph{Tll} is a strong repressor of gap genes and exhibits reduced expression in the absence of \emph{Zld}. Consequently, the ectopic expression of \emph{Kni} can be attributed to the delayed expression of \emph{Tll} caused by the lack of \emph{Zld}.
\end{itemize}

Since the only re-discovered third-order interaction involves ectopic expression rather than normal gene expression, we focus on second-order interactions and summarize all re-discovered interactions in Table \ref{t3}. 
\begin{table}[ht]
\renewcommand{\arraystretch}{0.5}
\begin{center}
\caption{\label{t3} TF interactions between \emph{Zld} and other TFs. \textbf{Rep.} represents the repetitions of interactions. $\overline{\textbf{PVIM}}$ represents the average value of PVIMs.}
\setlength{\tabcolsep}{4pt}
\begin{tabular}{c|ccc|cccc}
\toprule
  \textbf{Type}& \textbf{Pattern} (\emph{Zld}, $\cdot$)  &  \textbf{Rep.}  &  $\overline{\textbf{PVIM}}$   & \textbf{Pattern} ($\cdot$, \emph{Zld}) &  \textbf{Rep.}  &  $\overline{\textbf{PVIM}}$ \\
\midrule
Synergistic Interaction &(\emph{Zld}, \emph{Twi})     &    22  & 0.0323  & (\emph{Twi}, \emph{Zld})     &    33   & 0.0090 \\\midrule
&(\emph{Zld}, \emph{Kni})  &    69     & 0.0459  &(\emph{Kni}, \emph{Zld})     &    15  & 2e-16 \\
Hierarchical Interaction &(\emph{Zld}, \emph{Bcd})  &    6    & 0.0357  & (\emph{Bcd}, \emph{Zld})     &    48  & 1.047e-16 \\
\multirow{2}{*}{(GG TFs)}&(\emph{Zld}, \emph{Gt})      &    5    & 0.0204  & (\emph{Gt}, \emph{Zld})     &    27  & 1.115e-16 \\
&(\emph{Zld}, \emph{Kr})      &    11   & 0.0106  & (\emph{Kr}, \emph{Zld})     &    29  & 1.024e-16  \\
&(\emph{Zld}, \emph{Tll})     &    4    & 0.0131  & (\emph{Tll}, \emph{Zld})     &    29 & 8.134e-17\\
\midrule
Hierarchical Interaction & (\emph{Zld}, \emph{Ftz})     &    29   & 0.0345  & (\emph{Ftz}, \emph{Zld})     &    27  & 1.141e-16\\
(AP TFs)&(\emph{Zld}, \emph{Cad})     &    5    & 0.0206  & (\emph{Cad}, \emph{Zld})     &    14  & 9.614e-17  \\

\bottomrule
\end{tabular}
\end{center}
  \vspace{-3mm}
\end{table}
The following observations can be made from the table:
\begin{itemize}[left=0pt]
    \item Synergistic Interaction:  The table shows that TFs \emph{Zld} and \emph{Twi} have a Synergistic Interaction because both (\emph{Zld}, \emph{Twi}) and (\emph{Twi}, \emph{Zld}) have the average PVIMs significantly larger than 0. This discovery is also consistent with the background knowledge that \emph{Twi} is just temporally delayed, but later fully recovered by nc 14 in the absence of \emph{Zld} \citep{nien2011temporal}. 

    \item Hierarchical Interaction (gap gene expression TFs nested in \emph{Zld}): iKF selects five pairwise interactions between gap-gene expression TFs and \emph{Zld} in both directions. When \emph{Zld} splits the root node, its average PVIMs of five interactions are larger than 0.01, which is large given the evaluation criterion is misclassification rate. 
    Meanwhile, when a gap gene expression TF, say \emph{Bcd}, is the ``King'', interaction (\emph{Bcd}, \emph{Zld}) has an almost zero average PVIM. This pattern means all five lists belong to Hierarchical Interaction, and all five gap-gene expression TFs are nested in \emph{Zld}. 

    \item Hierarchical Interaction (anteroposterior patterning TFs Nested in \emph{Zld}): Similarly, the anteroposterior patterning TFs, \emph{Cad} and \emph{Ftz} are also nested in \emph{Zld}. The discovery of Hierarchical Interactions surrounding dominant TF \emph{Zld} is consistent with the previously introduced background knowledge. 
\end{itemize}

\begin{table}[ht]
\renewcommand{\arraystretch}{0.8}
\begin{center}
\caption{\label{t4} Kings, Kings' average PVIM and their roles in the mechanism.}
\scalebox{0.9}{
\begin{tabular}{cccccccc}
\toprule
King $\backslash$ Depth     & $d$=1    & $d$=2    & $d$=3    & $d$=4     & Role  \\
\midrule
\emph{Kni}      & -4.774e-17 & -3.525e-17 & 9.714e-19  & 2.220e-18   & Nested TF  \\
\emph{Twi}      & -2.318e-17 & 0.0217     & 0.00578    & 0.00589     & Dominant TF  \\
\emph{Zld}      & -4.163e-19 & 0.0296     & 0.0118     & 0.0155      & Dominant TF  \\
\emph{Bcd}      &  2.359e-18 & 3.747e-18  & -1.776e-17 & 1.207e-17   & Nested TF  \\
\emph{Cad}      &  1.388e-18 & 1.110e-17  & -6.106e-18 & -2.082e-18  & Nested TF   \\
\emph{Gt}       & -4.163e-18 & -4.580e-18 & 7.910e-18  & 4.996e-18   & Nested TF  \\
\emph{Kr}       & 1.568e-17  & -4.996e-18 & -1.429e-17 & 3.303e-17   & Nested TF \\
\emph{Tll}      & -5.829e-18 & 2.109e-17  & -1.388e-18 & -1.284e-17  & Nested TF  \\
\emph{Ftz}      & -4.857e-18 & -3.747e-18 & 1.207e-17  & -2.193e-17  & Nested TF   \\
\bottomrule
\end{tabular}
}
\end{center}
\end{table}

To further validate the interaction orders associated with the "Kings," Table \ref{t4} presents all nine TFs identified as "Kings" and tracks the changes in their Kings' PVIMs as the maximum depth $d$ increases. When $d=1$, all Kings' PVIMs are near zero, indicating the absence of main effects for these "Kings." At $d=2$, the Kings' PVIMs of \emph{Zld} and \emph{Twi} exceed 0.02, while those of the other seven TFs remain close to zero. This suggests that \emph{Zld} and \emph{Twi} participate in at least one Synergistic Interaction or act as the \emph{dominant TFs} in Hierarchical Interactions, whereas the other seven TFs can only function as \emph{nested TFs} in Hierarchical Interactions. As $d$ increases to 3 and 4, the Kings' PVIMs of \emph{Zld} and \emph{Twi} stabilize, and the Kings' PVIMs for the remaining TFs stay negligible, indicating no involvement in higher-order interactions (orders 3 or 4). Together, Tables \ref{t3} and \ref{t4} elucidate the type and order of interactions involving the "Kings," thereby providing key insights into the mechanisms governing early Drosophila embryogenesis.

%% file: section/conclusion.tex
\section{Conclusion}

In this paper, we introduce iKF, a method designed to reveal the underlying mechanisms of data through variable selection and interaction identification. The key contributions of iKF are three-fold. First, it enhances the traditional tree construction process by prioritizing critical variables, referred to as "Kings," to systematically identify and model interactions involving these variables. Second, iKF seamlessly integrates variable and interaction selection by iteratively assessing tree importance and updating variable weights. Third, it categorizes interactions into three distinct types and establishes criteria for determining their type and order. We demonstrate the effectiveness of iKF in identifying complex interactions through a series of experimental scenarios and a real-world scientific re-discovery case, highlighting its potential for advancing variable selection and interaction discovery across diverse scientific fields.


%% file: section/app.tex
\newpage
\section*{Appendix}

\section{Appendix for Scientific Discovery in Math}

\textbf{Simulated setting.} The main purpose of simulation studies is to assess the performance of the \emph{Iteratively Kings' Forests} (iKF) in feature screening and interaction identification through comparison with feature screening procedure DC-SIS \citep{li2012feature} and the Iterative Random Forest (iRF) \citep{basu2018iterative}. Especially, we focus on comparing them under the model settings that important variables affect the response through participating in some interaction effects. 

Simulation settings are divided into two parts. In the five settings of part \textbf{(a)}, four variables affect the response through two pairwise interactions of different function forms.
\begin{itemize}
  \item [\textbf{(a1):} ]$y=s*x_{1}*x_{3}-s*x_{5}*(x_{7}<0.2)+e$
  \item [\textbf{(a2):} ]$y=2*s*x_{1}*\sin(x_{3})+2*s*x_{5}*\cos(x_{7}+\pi/2)+e$
  \item [\textbf{(a3):} ]$y=s*\exp(x_{1})*x_{3}/2-s*\log(5*|x_{5}|)*x_{7}+e$
  \item [\textbf{(a4):} ]$y=s*x_{1}x_{3}^{2}/2-s*sign(x_{5})*x_{7}^{2}+e$
  \item [\textbf{(a5):} ]$y=s*x_{1}*x_{3}+1.5*s*x_{5}*\sin(x_{7})+e$
\end{itemize}
In the five cases of part \textbf{(b)}, three variables together affect the response through one third-order interaction.
\begin{itemize}
  \item [\textbf{(b1):} ]$y=s*x_{1}*(1+x_{3})^2*\sin(x_{5})+e$
  \item [\textbf{(b2):} ]$y=s*x_{1}*\log(5*|1+x_{3}|)*\sin(x_{5})+e$
  \item [\textbf{(b3):} ]$y=s*x_{1}*sign(1+x_{3})*\sin(x_{5})+e$
  \item [\textbf{(b4):} ]$y=s*x_{1}*x_{3}*\sin(x_{5})+e$
  \item [\textbf{(b5):} ]$y=s*x_{1}*x_{3}*x_{5}+e$
\end{itemize}


\subsection{Comparison with DC-SIS in Feature Screening}
In this part, we compare iKF's performance in feature screening with DC-SIS\citep{li2012feature}. 

\begin{table}[ht]
\renewcommand{\arraystretch}{0.8}
\begin{center}
\caption{\label{c41} Comparison between iKF and DC-SIS with respect to Quantiles of MRS.}
\scalebox{0.85}{
\begin{tabular}{c|ccccc|ccccc}
\toprule
MRS  & \multicolumn{5}{c|}{iKF}  & \multicolumn{5}{c}{DC-SIS} \\
\midrule
Quantile  & 5\%   & 25\%   & 50\%   & 75\%  & 95\%   & 5\%   & 25\%  & 50\%   & 75\%   & 95\% \\
\midrule
     \multicolumn{11}{c}{$n=200$ and $p=200$}\cr  
\midrule
$(a1)$  & 4  & 4  & 4   & 8   & 22   & 9   & 20  & 37   & 62   & 117  \\
$(a2)$  & 4  & 4  & 4   & 23  & 58   & 24  & 44  & 70   & 103  & 152  \\
$(a3)$  & 4  & 4  & 5   & 8   & 24   & 6   & 15  & 29   & 52   & 110  \\
$(a4)$  & 4  & 4  & 4   & 5   & 9    & 4   & 7   & 13   & 26   & 65   \\
$(a5)$  & 4  & 4  & 16  & 40  & 107  & 16  & 37  & 57   & 87   & 128  \\
\midrule
$(b1)$  & 3  & 3  & 3   & 4   & 23   & 4   & 9   & 24   & 49   & 124 \\
$(b2)$  & 3  & 3  & 5   & 11  & 38   & 5   & 10  & 20   & 35   & 91  \\
$(b3)$  & 4  & 8  & 23  & 50  & 112  & 28  & 65  & 122  & 155  & 187  \\
$(b4)$  & 4  & 9  & 16  & 36  & 80   & 20  & 40  & 64   & 87   & 129  \\
$(b5)$  & 3  & 4  & 6   & 9   & 18   & 6   & 10  & 18   & 28   & 67  \\
\midrule
     \multicolumn{11}{c}{$n=200$ and $p=500$}\cr  
\midrule
$(a1)$  & 4  & 4  & 12  & 28   & 93   & 18  & 49   & 89   & 150  & 326  \\
$(a2)$  & 4  & 4  & 18  & 80   & 196  & 49  & 115  & 177  & 263  & 383  \\
$(a3)$  & 4  & 5  & 9   & 17   & 56   & 9   & 32   & 66   & 129  & 281  \\
$(a4)$  & 4  & 4  & 5   & 11   & 25   & 6   & 13   & 25   & 53   & 142  \\
$(a5)$  & 4  & 8  & 48  & 110  & 244  & 41  & 91   & 160  & 223  & 326 \\
\midrule
$(b1)$  & 3   & 3   & 3    & 6    & 31   & 5   & 22  & 62   & 140  & 259 \\
$(b2)$  & 3   & 4   & 11   & 27   & 109  & 6   & 20  & 39   & 74   & 161  \\
$(b3)$  & 31  & 96  & 194  & 323  & 412  & 54  & 147 & 240  & 391  & 471 \\
$(b4)$  & 8   & 20  & 49   & 88   & 180  & 41  & 71  & 109  & 166  & 305 \\
$(b5)$  & 3   & 5   & 12   & 19   & 42   & 7   & 21  & 37   & 63   & 175  \\
\bottomrule
\end{tabular}
}
\end{center}
\end{table}

Table \ref{c41} shows that iKF outperforms DC-SIS in all cases with respect to minimum recovery size (MRS). When $p=200$, iKF ranks all four important variables in settings (a1)-(a4) as the top five in more than 50\% repetitions. Meanwhile, DC-SIS gives
MRS larger than 37 in more than 75\% of times for cases (a2), (a5), (b3) and (b4), which means that DC-SIS will miss at least one important variable most of the time under these settings. When we increase $p$ to 500, iKF still outperforms DC-SIS a lot in all settings with respect to MRS, and works well in feature screening for all cases except for (b3).  

\begin{table}[ht]
\renewcommand{\arraystretch}{0.8}
\begin{center}
\caption{\label{c42} Comparison between iKF and DC-SIS for ($\textbf{a}$) Settings}\vspace{0.2cm}
\scalebox{0.9}{
\begin{tabular}{cc|cccc|c|cccc|c}
\toprule
 & & \multicolumn{5}{c|}{iKF}  & \multicolumn{5}{c}{DC-SIS} \\
 \midrule
& & \multicolumn{4}{c|}{$P_{s}$} & $P_{a}$  & \multicolumn{4}{c|}{$P_{s}$} & $P_{a}$ \\
      \midrule
 Model & Size & $X_{1}$ & $X_{3}$ & $X_{5}$ & $X_{7}$ & all  & $X_{1}$ & $X_{3}$ & $X_{5}$ & $X_{7}$ & all \\
\midrule
    \multicolumn{11}{c}{$n=200$ and $p=200$}\cr  
\midrule
$(a1)$ & $d_{1}$  & 0.98 & 0.97 & 1.00 & 0.96  & 0.92 & 0.68 & 0.67 & 1.00 & 0.46 & 0.22 \\
       & $d_{2}$  & 0.99 & 1.00 & 1.00 & 0.99  & 0.98 & 0.87 & 0.87 & 1.00 & 0.67 & 0.50 \\
$(a2)$ & $d_{1}$  & 0.97 & 0.86 & 0.93 & 0.87  & 0.71 & 0.70 & 0.25 & 0.68 & 0.26 & 0.02 \\
       & $d_{2}$  & 0.99 & 0.94 & 0.99 & 0.92  & 0.86 & 0.87 & 0.47 & 0.88 & 0.49 & 0.18 \\
$(a3)$ & $d_{1}$  & 0.99 & 1.00 & 0.94 & 1.00  & 0.93 & 0.78 & 1.00 & 0.39 & 1.00 & 0.27 \\
       & $d_{2}$  & 0.99 & 1.00 & 0.99 & 1.00  & 0.98 & 0.89 & 1.00 & 0.63 & 1.00 & 0.56 \\
$(a4)$ & $d_{1}$  & 1.00 & 0.98 & 1.00 & 1.00  & 0.98 & 1.00 & 0.61 & 1.00 & 0.99 & 0.60 \\
       & $d_{2}$  & 1.00 & 0.99 & 1.00 & 1.00  & 0.99 & 1.00 & 0.84 & 1.00 & 1.00 & 0.84 \\
$(a5)$ & $d_{1}$  & 0.87 & 0.89 & 0.87 & 0.69  & 0.53 & 0.50 & 0.55 & 0.64 & 0.27 & 0.03 \\
       & $d_{2}$  & 0.94 & 0.97 & 0.95 & 0.82  & 0.73 & 0.78 & 0.81 & 0.91 & 0.43 & 0.21 \\
\midrule
      \multicolumn{11}{c}{$n=200$ and $p=500$}\cr  
\midrule
$(a1)$ & $d_{1}$  & 0.83 & 0.83 & 1.00 & 0.83  & 0.62 & 0.38 & 0.41 & 1.00 & 0.27 & 0.05 \\
       & $d_{2}$  & 0.92 & 0.95 & 1.00 & 0.95  & 0.83 & 0.60 & 0.62 & 1.00 & 0.43 & 0.16 \\
$(a2)$ & $d_{1}$  & 0.82 & 0.64 & 0.87 & 0.77  & 0.51 & 0.36 & 0.10 & 0.34 & 0.11 & 0.01 \\
       & $d_{2}$  & 0.87 & 0.75 & 0.95 & 0.84  & 0.64 & 0.58 & 0.19 & 0.53 & 0.21 & 0.03 \\
$(a3)$ & $d_{1}$  & 0.97 & 1.00 & 0.80 & 1.00  & 0.77 & 0.67 & 1.00 & 0.17 & 1.00 & 0.14 \\
       & $d_{2}$  & 0.99 & 1.00 & 0.90 & 1.00  & 0.89 & 0.77 & 1.00 & 0.31 & 1.00 & 0.26 \\
$(a4)$ & $d_{1}$  & 1.00 & 0.89 & 1.00 & 1.00  & 0.89 & 1.00 & 0.40 & 1.00 & 0.93 & 0.38 \\
       & $d_{2}$  & 1.00 & 0.97 & 1.00 & 1.00  & 0.97 & 1.00 & 0.59 & 1.00 & 0.99 & 0.59 \\
$(a5)$ & $d_{1}$  & 0.67 & 0.73 & 0.72 & 0.54  & 0.31 & 0.29 & 0.22 & 0.52 & 0.12 & 0.00 \\
       & $d_{2}$  & 0.82 & 0.80 & 0.88 & 0.61  & 0.43 & 0.48 & 0.42 & 0.78 & 0.21 & 0.02 \\
\midrule
\end{tabular}
}
\end{center}
\end{table}

\begin{table}[ht]
\renewcommand{\arraystretch}{0.8}
\begin{center}
\caption{\label{c43} Comparison between iKF and DC-SIS for ($\textbf{b}$) Settings}\vspace{0.2cm}
\scalebox{0.9}{
\begin{tabular}{cc|ccc|c|ccc|c}
  \midrule
 & & \multicolumn{4}{c|}{iKF}  & \multicolumn{4}{c}{DC-SIS} \\
 \midrule
& & \multicolumn{3}{c|}{$P_{s}$} & $P_{a}$  & \multicolumn{3}{c|}{$P_{s}$} & $P_{a}$ \\
      \midrule
 Model & Size & $X_{1}$ & $X_{3}$ & $X_{5}$ & all  & $X_{1}$ & $X_{3}$ & $X_{5}$ & all \\
\midrule
\multicolumn{9}{c}{$n=200$ and $p=200$}\cr  
\midrule
$(b1)$ & $d_{1}$  & 1.00 & 1.00 & 0.93 & 0.93 & 0.92 & 1.00 & 0.45 & 0.44 \\
       & $d_{2}$  & 1.00 & 1.00 & 0.97 & 0.97 & 0.99 & 1.00 & 0.66 & 0.66 \\
$(b2)$ & $d_{1}$  & 1.00 & 0.87 & 1.00 & 0.87 & 0.99 & 0.76 & 0.62 & 0.49 \\
       & $d_{2}$  & 1.00 & 0.95 & 1.00 & 0.95 & 1.00 & 0.90 & 0.87 & 0.79 \\
$(b3)$ & $d_{1}$  & 1.00 & 0.47 & 1.00 & 0.47 & 0.94 & 0.06 & 0.37 & 0.02 \\
       & $d_{2}$  & 1.00 & 0.65 & 1.00 & 0.65 & 1.00 & 0.16 & 0.56 & 0.11 \\
$(b4)$ & $d_{1}$  & 0.96 & 0.97 & 0.63 & 0.60 & 0.62 & 0.62 & 0.22 & 0.05 \\
       & $d_{2}$  & 0.98 & 0.99 & 0.79 & 0.77 & 0.85 & 0.77 & 0.38 & 0.20 \\
$(b5)$ & $d_{1}$  & 0.99 & 0.98 & 0.99 & 0.96 & 0.80 & 0.76 & 0.79 & 0.51 \\
       & $d_{2}$  & 1.00 & 0.99 & 1.00 & 0.99 & 0.94 & 0.90 & 0.94 & 0.82 \\
\midrule
\multicolumn{9}{c}{$n=200$ and $p=500$}\cr  
\midrule
$(b1)$ & $d_{1}$  & 0.94 & 1.00 & 0.92 & 0.90 & 0.77 & 1.00 & 0.26 & 0.24 \\
       & $d_{2}$  & 0.97 & 1.00 & 0.97 & 0.96 & 0.91 & 1.00 & 0.42 & 0.39 \\
$(b2)$ & $d_{1}$  & 1.00 & 0.67 & 1.00 & 0.67 & 0.93 & 0.67 & 0.35 & 0.22 \\
       & $d_{2}$  & 1.00 & 0.81 & 1.00 & 0.81 & 0.99 & 0.86 & 0.56 & 0.48 \\
$(b3)$ & $d_{1}$  & 1.00 & 0.02 & 0.88 & 0.02 & 0.73 & 0.04 & 0.10 & 0.00 \\
       & $d_{2}$  & 1.00 & 0.08 & 0.94 & 0.08 & 0.94 & 0.05 & 0.30 & 0.04 \\
$(b4)$ & $d_{1}$  & 0.83 & 0.88 & 0.29 & 0.23 & 0.47 & 0.43 & 0.08 & 0.01 \\
       & $d_{2}$  & 0.96 & 0.94 & 0.54 & 0.46 & 0.64 & 0.61 & 0.14 & 0.04 \\
$(b5)$ & $d_{1}$  & 0.91 & 0.93 & 0.86 & 0.72 & 0.61 & 0.59 & 0.54 & 0.21 \\
       & $d_{2}$  & 0.97 & 1.00 & 0.97 & 0.93 & 0.86 & 0.77 & 0.77 & 0.51 \\
\midrule
\end{tabular}
}
\end{center}
\end{table}

Tables \ref{c42}-\ref{c43} give the proportions that active predictors are selected for given model sizes. iKF again outpeforms DC-SIS in all cases. Especially, DC-SIS fails to include all active predictors in cases (a2), (a5), (b3) and (b4), while iKF still works pretty well. Furthermore, as $p$ increases from 200 to 500, iKF has even larger advantage over DC-SIS with respect to $P_{a}$. If we consider $p=500$ and the model size $d_{2}=[n/\log(n)]$ for comparison, iKF gives $P_{a}$ larger than 80\% in most cases, while DC-SIS gives $P_{a}$ less than 60\% in all situations. Therefore, iKF performs much better than DC-SIS in feature screening when the active variables have no main effect, and affect the response only through participating in interaction effects.

\newpage
\subsection{Comparison with Iterative Random Forest}


\begin{table}[ht]
\renewcommand{\arraystretch}{0.8}
\begin{center}
\caption{\label{c44} Quantiles of MRS for iKF and iRF}\vspace{0.2cm}
\scalebox{0.85}{
\begin{tabular}{c|ccccc|ccccc}
\midrule
MRS  & \multicolumn{5}{c|}{iKF}  & \multicolumn{5}{c}{iRF} \\
\midrule
Quantile  & 5\%   & 25\%   & 50\%   & 75\%  & 95\%   & 5\%   & 25\%  & 50\%   & 75\%   & 95\% \\
\midrule
     \multicolumn{11}{c}{$n=200$ and $p=200$}\cr  
\midrule
$(a1)$  & 4  & 4  & 4   & 8   & 22  & 4  & 5  & 6  & 8   & 27  \\
$(a2)$  & 4  & 4  & 4   & 23  & 58  & 8  & 13 & 23 & 43  & 84 \\
$(a3)$  & 4  & 4  & 5   & 8   & 24  & 4  & 5  & 7  & 10   & 21 \\
$(a4)$  & 4  & 4  & 4   & 5   & 9   & 4  & 4  & 4  & 5   & 11 \\
$(a5)$  & 4  & 4  & 16  & 40  & 107 & 9  & 14 & 25 & 54  & 141 \\
\midrule
$(b1)$  & 3  & 3  & 3   & 4   & 23  & 11 & 18 & 35 & 58  & 120 \\
$(b2)$  & 3  & 3  & 5   & 11  & 38  & 5  & 8  & 14 & 22  & 52 \\
$(b3)$  & 4  & 8  & 23  & 50  & 112 & 13 & 61 & 123 & 149 & 157 \\
$(b4)$  & 4  & 9  & 16  & 36  & 80  & 6  & 20 & 39 & 61  & 126 \\
$(b5)$  & 3  & 4  & 6   & 9   & 18  & 5  & 8  & 13 & 21  & 53 \\
\midrule
     \multicolumn{11}{c}{$n=200$ and $p=500$}\cr  
\midrule
$(a1)$  & 4  & 4  & 12  & 28   & 93  & 5  & 8  & 12 & 27  & 120 \\
$(a2)$  & 4  & 4  & 18  & 80   & 196 & 16 & 48 & 85 & 138 & 266 \\
$(a3)$  & 4  & 5  & 9   & 17   & 56  & 4  & 7  & 12 & 21  & 68   \\
$(a4)$  & 4  & 4  & 5   & 11   & 25  & 4  & 4  & 5  & 7   & 16  \\
$(a5)$  & 4  & 8  & 48  & 110  & 244 & 10 & 30 & 61 & 104 & 260  \\
\midrule
$(b1)$  & 3   & 3   & 3    & 6    & 31  & 15  & 42   & 76   & 140  & 225  \\
$(b2)$  & 3   & 4   & 11   & 27   & 109 & 10  & 20   & 39   & 73   & 235  \\
$(b3)$  & 31  & 96  & 194  & 323  & 412 & 37  & 141  & 248  & 257  & 266 \\
$(b4)$  & 8   & 20  & 49   & 88   & 180 & 15  & 34   & 77   & 143  & 255  \\
$(b5)$  & 3   & 5   & 12   & 19   & 42  & 6   & 14   & 26   & 52   & 123 \\
\midrule
\end{tabular}
}
\end{center}
\end{table}

Table \ref{c44} shows that iKF outperforms iRF in most cases with respect to minimum model size. However, its advantages over iRF is not as large as the advantages over DC-SIS. Among all ten settings, (b3) is the most difficult for both algorithms to discover. When we increase $p$ to 500, iKF still ranks all important variables as the top 20 in more than 50\% repetitions for the cases (a1)-(a4), (b1), (b2) and (b5). At the same time, iRF could only achieve it in the cases (a1), (a3) and (a4).

\begin{table}[ht]
\renewcommand{\arraystretch}{0.8}
\begin{center}
\caption{\label{c5} Comparison between iKF and iRF for $\textbf{a}$ Settings}\vspace{0.2cm}
\scalebox{0.9}{
\begin{tabular}{cc|cccc|c|cccc|c}
  \midrule
 & & \multicolumn{5}{c|}{iKF}  & \multicolumn{5}{c}{iRF} \\
 \midrule
& & \multicolumn{4}{c|}{$P_{s}$} & $P_{a}$  & \multicolumn{4}{c|}{$P_{s}$} & $P_{a}$ \\
      \midrule
 Model & Size & $X_{1}$ & $X_{3}$ & $X_{5}$ & $X_{7}$ & all  & $X_{1}$ & $X_{3}$ & $X_{5}$ & $X_{7}$ & all \\
\midrule
    \multicolumn{11}{c}{$n=200$ and $p=200$}\cr  
\midrule
$(a1)$ & $d_{1}$  & 0.98 & 0.97 & 1.00 & 0.96  & 0.92 & 0.98 & 0.98 & 1.00 & 0.94 & 0.91 \\
       & $d_{2}$  & 0.99 & 1.00 & 1.00 & 0.99  & 0.98 & 0.99 & 1.00 & 1.00 & 0.98 & 0.97 \\
$(a2)$ & $d_{1}$  & 0.97 & 0.86 & 0.93 & 0.87  & 0.71 & 0.96 & 0.59 & 0.99 & 0.71 & 0.40 \\
       & $d_{2}$  & 0.99 & 0.94 & 0.99 & 0.92  & 0.86 & 1.00 & 0.81 & 1.00 & 0.83 & 0.68 \\
$(a3)$ & $d_{1}$  & 0.99 & 1.00 & 0.94 & 1.00  & 0.93 & 0.99 & 1.00 & 0.94 & 1.00 & 0.93 \\
       & $d_{2}$  & 0.99 & 1.00 & 0.99 & 1.00  & 0.98 & 1.00 & 1.00 & 0.99 & 1.00 & 0.99 \\
$(a4)$ & $d_{1}$  & 1.00 & 0.98 & 1.00 & 1.00  & 0.98 & 1.00 & 0.97 & 1.00 & 1.00 & 0.97 \\
       & $d_{2}$  & 1.00 & 0.99 & 1.00 & 1.00  & 0.99 & 1.00 & 0.99 & 1.00 & 1.00 & 0.99 \\
$(a5)$ & $d_{1}$  & 0.87 & 0.89 & 0.87 & 0.69  & 0.53 & 0.87 & 0.83 & 0.95 & 0.50 & 0.37 \\
       & $d_{2}$  & 0.94 & 0.97 & 0.95 & 0.82  & 0.73 & 0.96 & 0.94 & 0.99 & 0.69 & 0.60 \\
\midrule
      \multicolumn{11}{c}{$n=200$ and $p=500$}\cr  
\midrule
$(a1)$ & $d_{1}$  & 0.83 & 0.83 & 1.00 & 0.83  & 0.62 & 0.91 & 0.90 & 1.00 & 0.87 & 0.70 \\
       & $d_{2}$  & 0.92 & 0.95 & 1.00 & 0.95  & 0.83 & 0.96 & 0.97 & 1.00 & 0.90 & 0.84 \\
$(a2)$ & $d_{1}$  & 0.82 & 0.64 & 0.87 & 0.77  & 0.51 & 0.89 & 0.30 & 0.80 & 0.25 & 0.07 \\
       & $d_{2}$  & 0.87 & 0.75 & 0.95 & 0.84  & 0.64 & 0.97 & 0.49 & 0.94 & 0.41 & 0.17  \\
$(a3)$ & $d_{1}$  & 0.97 & 1.00 & 0.80 & 1.00  & 0.77 & 1.00 & 1.00 & 0.72 & 1.00 & 0.72 \\
       & $d_{2}$  & 0.99 & 1.00 & 0.90 & 1.00  & 0.89 & 1.00 & 1.00 & 0.88 & 1.00 & 0.88 \\
$(a4)$ & $d_{1}$  & 1.00 & 0.89 & 1.00 & 1.00  & 0.89 & 1.00 & 0.96 & 1.00 & 1.00 & 0.96 \\
       & $d_{2}$  & 1.00 & 0.97 & 1.00 & 1.00  & 0.97 & 1.00 & 1.00 & 1.00 & 1.00 & 1.00 \\
$(a5)$ & $d_{1}$  & 0.67 & 0.73 & 0.72 & 0.54  & 0.31 & 0.64 & 0.74 & 0.83 & 0.25 & 0.11 \\
       & $d_{2}$  & 0.82 & 0.80 & 0.88 & 0.61  & 0.43 & 0.80 & 0.85 & 0.94 & 0.49 & 0.33\\
\midrule
\end{tabular}
}
\end{center}
\end{table}

\begin{table}[ht]
\renewcommand{\arraystretch}{0.8}
\begin{center}
\caption{\label{c6} Comparison between iKF and iRF for $\textbf{b}$ Settings}\vspace{0.2cm}
\scalebox{0.9}{
\begin{tabular}{cc|ccc|c|ccc|c}
  \midrule
 & & \multicolumn{4}{c|}{iKF}  & \multicolumn{4}{c}{iRF} \\
 \midrule
& & \multicolumn{3}{c|}{$P_{s}$} & $P_{a}$  & \multicolumn{3}{c|}{$P_{s}$} & $P_{a}$ \\
      \midrule
 Model & Size & $X_{1}$ & $X_{3}$ & $X_{5}$ & all  & $X_{1}$ & $X_{3}$ & $X_{5}$ & all \\
\midrule
\multicolumn{9}{c}{$n=200$ and $p=200$}\cr  
\midrule
$(b1)$ & $d_{1}$  & 1.00 & 1.00 & 0.93 & 0.93 & 0.80 & 0.99 & 0.29 & 0.26 \\
       & $d_{2}$  & 1.00 & 1.00 & 0.97 & 0.97 & 0.95 & 1.00 & 0.58 & 0.57 \\
$(b2)$ & $d_{1}$  & 1.00 & 0.87 & 1.00 & 0.87 & 1.00 & 0.73 & 0.88 & 0.64 \\
       & $d_{2}$  & 1.00 & 0.95 & 1.00 & 0.95 & 1.00 & 0.93 & 0.97 & 0.90 \\
$(b3)$ & $d_{1}$  & 1.00 & 0.47 & 1.00 & 0.47 & 1.00 & 0.07 & 0.98 & 0.07 \\
       & $d_{2}$  & 1.00 & 0.65 & 1.00 & 0.65 & 1.00 & 0.12 & 1.00 & 0.12 \\
$(b4)$ & $d_{1}$  & 0.96 & 0.97 & 0.63 & 0.60 & 0.88 & 0.90 & 0.29 & 0.22 \\
       & $d_{2}$  & 0.98 & 0.99 & 0.79 & 0.77 & 0.94 & 0.97 & 0.51 & 0.47 \\
$(b5)$ & $d_{1}$  & 0.99 & 0.98 & 0.99 & 0.96 & 0.85 & 0.89 & 0.90 & 0.68 \\
       & $d_{2}$  & 1.00 & 0.99 & 1.00 & 0.99 & 0.97 & 0.96 & 0.96 & 0.89 \\
\midrule
\multicolumn{9}{c}{$n=200$ and $p=500$}\cr  
\midrule
$(b1)$ & $d_{1}$  & 0.94 & 1.00 & 0.92 & 0.90 & 0.62 & 0.98 & 0.09 & 0.07 \\
       & $d_{2}$  & 0.97 & 1.00 & 0.97 & 0.96 & 0.79 & 1.00 & 0.28 & 0.24 \\
$(b2)$ & $d_{1}$  & 1.00 & 0.67 & 1.00 & 0.67 & 1.00 & 0.41 & 0.48 & 0.24 \\
       & $d_{2}$  & 1.00 & 0.81 & 1.00 & 0.81 & 1.00 & 0.60 & 0.73 & 0.48 \\
$(b3)$ & $d_{1}$  & 1.00 & 0.02 & 0.88 & 0.02 & 1.00 & 0.04 & 0.74 & 0.02 \\
       & $d_{2}$  & 1.00 & 0.08 & 0.94 & 0.08 & 1.00 & 0.06 & 0.89 & 0.06 \\
$(b4)$ & $d_{1}$  & 0.83 & 0.88 & 0.29 & 0.23 & 0.75 & 0.83 & 0.14 & 0.07 \\
       & $d_{2}$  & 0.96 & 0.94 & 0.54 & 0.46 & 0.90 & 0.95 & 0.32 & 0.28 \\
$(b5)$ & $d_{1}$  & 0.91 & 0.93 & 0.86 & 0.72 & 0.74 & 0.72 & 0.71 & 0.35 \\
       & $d_{2}$  & 0.97 & 1.00 & 0.97 & 0.93 & 0.87 & 0.86 & 0.86 & 0.62 \\
\midrule
\end{tabular}
}
\end{center}
\end{table}

Tables \ref{c5}-\ref{c6} give the proportions that active predictors are selected for given model sizes $d_{1}$ and $d_{2}$. iKF outpeforms iRF in all \textbf{(b)} settings. For the settings of \textbf{(a)}, iKF wins the cases (a2) and (a5), while iRF performs slightly better in (a1) and (a4). In general, iRF performs much better than DC-SIS in selecting participants of active interaction effects, but still not as good as iKF.

\begin{table}[ht]
\renewcommand{\arraystretch}{0.8}
\begin{center}
\caption{\label{c7} Successful Selecting Rate of Interactions}\vspace{0.2cm}
\scalebox{0.85}{
\begin{tabular}{c|cc|c|c|c|c|cc|c|c|c}
\midrule
\multicolumn{6}{c|}{iKF}  & \multicolumn{6}{c}{iRF} \\
\midrule
Model & \multicolumn{2}{c|}{$p_{inter}$} & $P_{inter}$ & Model & $P_{inter}$ & Model  & \multicolumn{2}{c|}{$p_{inter}$}   &$P_{inter}$ & Model & $P_{inter}$ \\
\midrule
\multicolumn{12}{c}{$n=200$ and $p=200$}\cr  
\midrule
$(a1)$  & 0.77 & 0.92 & 0.71 & $(b1)$ & 0.87 & $(a1)$  & 0.34 & 0.94 & 0.32 & $(b1)$ & 0.08 \\
$(a2)$  & 0.76 & 0.75 & 0.58 & $(b2)$ & 0.67 & $(a2)$  & 0.46 & 0.48 & 0.16 & $(b2)$ & 0.01 \\
$(a3)$  & 0.98 & 0.76 & 0.74 & $(b3)$ & 0.53 & $(a3)$  & 0.65 & 0.90 & 0.56 & $(b3)$ & 0    \\
$(a4)$  & 0.93 & 0.99 & 0.92 & $(b4)$ & 0.15 & $(a4)$  & 0.82 & 0.99 & 0.81 & $(b4)$ & 0    \\
$(a5)$  & 0.64 & 0.57 & 0.34 & $(b5)$ & 0.21 & $(a5)$  & 0.53 & 0.33 & 0.13 & $(b5)$ & 0.09 \\
\midrule
\multicolumn{12}{c}{$n=200$ and $p=500$}\cr  
\midrule
$(a1)$  & 0.73 & 0.82 & 0.60 & $(b1)$ & 0.68  & $(a1)$  & 0.12 & 0.84 & 0.09 & $(b1)$ & 0  \\ 
$(a2)$  & 0.53 & 0.73 & 0.41 & $(b2)$ & 0.48  & $(a2)$  & 0.19 & 0.15 & 0.02 & $(b2)$ & 0  \\
$(a3)$  & 0.92 & 0.54 & 0.51 & $(b3)$ & 0.16  & $(a3)$  & 0.62 & 0.76 & 0.45 & $(b3)$ & 0  \\
$(a4)$  & 0.80 & 0.97 & 0.77 & $(b4)$ & 0.03  & $(a4)$  & 0.65 & 0.96 & 0.61 & $(b4)$ & 0  \\
$(a5)$  & 0.57 & 0.45 & 0.28 & $(b5)$ & 0.05  & $(a5)$  & 0.28 & 0.14 & 0.04 & $(b5)$ & 0  \\
\midrule
\end{tabular}
}
\end{center}
\end{table}

Tables \ref{c7} gives the successful selecting rate of interactions for both procedures. iKF outperforms iRF for all cases. Especially for \textbf{(b)} settings, iRF totally fails to identify the third order interactions, while iKF has more than 50\% probability to discover them in cases (b1)-(b3) when $p=200$. For \textbf{(a)} settings, iKF also has a large advantage over iRF. Since iRF gives a list of all candidate interactions with non-zero stability score, the list is much longer than what is given by iKF in all settings. Therefore, iKF is not only more accurate, but also more efficient than iRF in discovering important interaction effects.

To conclude, iKF outperforms iRF in discovering both the interaction effects and the marginal unimportant variables that participates in these interaction effects.

\newpage
~\\
\newpage
~\\
\newpage
\section{Appendix for Scientific Discovery in Biology}
We list the detailed reference evidence for the identified interaction in Table~\ref{tab:inter_irf_ikf} and Table~\ref{tab:inter_ikf}.

\begin{table}[ht]
\renewcommand{\arraystretch}{0.8}
\begin{center}
\caption{\label{c10} Pairwise TF Interactions Recovered by Both iRF and iKF}\vspace{0.2cm}
\label{tab:inter_irf_ikf}
\scalebox{1}{
\begin{tabular}{cccccc}
\toprule
 Interaction & References \\
\midrule
 (Gt, Zld)  & \cite{harrison2011zelda}; \cite{nien2011temporal} \\
 (Twi, Zld) & \cite{harrison2011zelda}; \cite{nien2011temporal} \\
 (Gt, Kr)   & \cite{kraut1991spatial}; \cite{struhl1992control}; \\
            & \cite{capovilla1992giant}; \cite{schulz1994autonomous}   \\
 (Gt, Twi)  & \cite{li2008transcription}       \\
 (Kr, Twi)  & \cite{li2008transcription}        \\
 (Kr, Zld)  & \cite{harrison2011zelda}; \cite{nien2011temporal}  \\
 (Bcd, Gt)  & \cite{kraut1991spatial}; \cite{eldon1991interactions}  \\
 (Bcd, Twi) & \cite{li2008transcription}      \\
 (Hb, Twi)  & \cite{zeitlinger2007whole}       \\
 (Med, Twi) & \cite{nguyen1998drosophila}      \\
 (Med, Zld) & \cite{harrison2011zelda}   \\
 (Hb, Zld)  & \cite{harrison2011zelda}; \cite{nien2011temporal} \\
 (Bcd, Kr)  & \cite{hoch1991gene}; \cite{hoch1990cis}   \\
 (Bcd, Zld) & \cite{harrison2011zelda}; \cite{nien2011temporal} \\
 (D, Twi)   &  - \\
 (Gt, Med)  &  - \\
\bottomrule
\end{tabular}
}
\end{center}
\end{table}

\begin{table}[ht]
\renewcommand{\arraystretch}{0.8}
\begin{center}
\caption{\label{c11} TF Interactions Recovered only by iKF}\vspace{0.2cm}
\label{tab:inter_ikf}
\scalebox{1}{
\begin{tabular}{cccccc}
\toprule
 Interaction (S) & References \\
\midrule
 (Kni, Zld)       & \cite{harrison2011zelda}; \cite{nien2011temporal} \\
 (Ftz, Zld)       & \cite{harrison2011zelda}; \cite{nien2011temporal} \\
 (Cad, Zld)       & \cite{harrison2011zelda}; \cite{nien2011temporal} \\
 (Zld, Kni, Tll)  & \cite{nien2011temporal}; \cite{moran2006tailless} \\
\bottomrule
\end{tabular}
}
\end{center}
\end{table}









%% file: main.bbl
\begin{thebibliography}{}

\bibitem[\protect\citeauthoryear{Amaratunga, Cabrera, and Lee}{Amaratunga et~al.}{2008}]{amaratunga2008enriched}
Amaratunga, D., J.~Cabrera, and Y.-S. Lee (2008).
\newblock Enriched random forests.
\newblock {\em Bioinformatics\/}~{\em 24\/}(18), 2010--2014.

\bibitem[\protect\citeauthoryear{Ancona, Ceolini, {\"O}ztireli, and Gross}{Ancona et~al.}{2017}]{ancona2017towards}
Ancona, M., E.~Ceolini, C.~{\"O}ztireli, and M.~Gross (2017).
\newblock Towards better understanding of gradient-based attribution methods for deep neural networks.
\newblock {\em arXiv preprint arXiv:1711.06104\/}.

\bibitem[\protect\citeauthoryear{Archer and Kimes}{Archer and Kimes}{2008}]{archer2008empirical}
Archer, K.~J. and R.~V. Kimes (2008).
\newblock Empirical characterization of random forest variable importance measures.
\newblock {\em Computational Statistics \& Data Analysis\/}~{\em 52\/}(4), 2249--2260.

\bibitem[\protect\citeauthoryear{Basu, Kumbier, Brown, and Yu}{Basu et~al.}{2018}]{basu2018iterative}
Basu, S., K.~Kumbier, J.~B. Brown, and B.~Yu (2018).
\newblock Iterative random forests to discover predictive and stable high-order interactions.
\newblock {\em Proceedings of the National Academy of Sciences\/}, 201711236.

\bibitem[\protect\citeauthoryear{Breiman}{Breiman}{2001}]{breiman2001random}
Breiman, L. (2001).
\newblock Random forests.
\newblock {\em Machine learning\/}~{\em 45\/}(1), 5--32.

\bibitem[\protect\citeauthoryear{Capovilla, Eldon, and Pirrotta}{Capovilla et~al.}{1992}]{capovilla1992giant}
Capovilla, M., E.~D. Eldon, and V.~Pirrotta (1992).
\newblock The giant gene of drosophila encodes a b-zip dna-binding protein that regulates the expression of other segmentation gap genes.
\newblock {\em Development\/}~{\em 114\/}(1), 99--112.

\bibitem[\protect\citeauthoryear{D{\'\i}az-Uriarte and De~Andres}{D{\'\i}az-Uriarte and De~Andres}{2006}]{diaz2006gene}
D{\'\i}az-Uriarte, R. and S.~A. De~Andres (2006).
\newblock Gene selection and classification of microarray data using random forest.
\newblock {\em BMC bioinformatics\/}~{\em 7\/}(1), 3.

\bibitem[\protect\citeauthoryear{Eldon and Pirrotta}{Eldon and Pirrotta}{1991}]{eldon1991interactions}
Eldon, E.~D. and V.~Pirrotta (1991).
\newblock Interactions of the drosophila gap gene giant with maternal and zygotic pattern-forming genes.
\newblock {\em Development\/}~{\em 111\/}(2), 367--378.

\bibitem[\protect\citeauthoryear{Elmarakeby, Hwang, Arafeh, Crowdis, Gang, Liu, AlDubayan, Salari, Kregel, Richter, et~al.}{Elmarakeby et~al.}{2021}]{elmarakeby2021biologically}
Elmarakeby, H.~A., J.~Hwang, R.~Arafeh, J.~Crowdis, S.~Gang, D.~Liu, S.~H. AlDubayan, K.~Salari, S.~Kregel, C.~Richter, et~al. (2021).
\newblock Biologically informed deep neural network for prostate cancer discovery.
\newblock {\em Nature\/}~{\em 598\/}(7880), 348--352.

\bibitem[\protect\citeauthoryear{Fan and Li}{Fan and Li}{2001}]{fan2001variable}
Fan, J. and R.~Li (2001).
\newblock Variable selection via nonconcave penalized likelihood and its oracle properties.
\newblock {\em Journal of the American statistical Association\/}~{\em 96\/}(456), 1348--1360.

\bibitem[\protect\citeauthoryear{Fan and Lv}{Fan and Lv}{2008}]{fan2008sure}
Fan, J. and J.~Lv (2008).
\newblock Sure independence screening for ultrahigh dimensional feature space.
\newblock {\em Journal of the Royal Statistical Society: Series B (Statistical Methodology)\/}~{\em 70\/}(5), 849--911.

\bibitem[\protect\citeauthoryear{Harrison, Li, Kaplan, Botchan, and Eisen}{Harrison et~al.}{2011}]{harrison2011zelda}
Harrison, M.~M., X.-Y. Li, T.~Kaplan, M.~R. Botchan, and M.~B. Eisen (2011).
\newblock Zelda binding in the early drosophila melanogaster embryo marks regions subsequently activated at the maternal-to-zygotic transition.
\newblock {\em PLoS genetics\/}~{\em 7\/}(10), e1002266.

\bibitem[\protect\citeauthoryear{Hoch, Schr{\"o}der, Seifert, and J{\"a}ckle}{Hoch et~al.}{1990}]{hoch1990cis}
Hoch, M., C.~Schr{\"o}der, E.~Seifert, and H.~J{\"a}ckle (1990).
\newblock cis-acting control elements for kr{\"u}ppel expression in the drosophila embryo.
\newblock {\em The EMBO journal\/}~{\em 9\/}(8), 2587--2595.

\bibitem[\protect\citeauthoryear{Hoch, Seifert, and J{\"a}ckle}{Hoch et~al.}{1991}]{hoch1991gene}
Hoch, M., E.~Seifert, and H.~J{\"a}ckle (1991).
\newblock Gene expression mediated by cis-acting sequences of the kr{\"u}ppel gene in response to the drosophila morphogens bicoid and hunchback.
\newblock {\em The EMBO journal\/}~{\em 10\/}(8), 2267--2278.

\bibitem[\protect\citeauthoryear{Hou, Ghashami, Kuznetsov, and Torkamani}{Hou et~al.}{2024}]{hou2024hlogformer}
Hou, Z., M.~Ghashami, M.~Kuznetsov, and M.~Torkamani (2024).
\newblock Hlogformer: A hierarchical transformer for representing log data.
\newblock {\em arXiv preprint arXiv:2408.16803\/}.

\bibitem[\protect\citeauthoryear{Hou, Leng, Yu, Xia, and Wu}{Hou et~al.}{2023}]{hou2023pathexpsurv}
Hou, Z., J.~Leng, J.~Yu, Z.~Xia, and L.-Y. Wu (2023).
\newblock Pathexpsurv: pathway expansion for explainable survival analysis and disease gene discovery.
\newblock {\em BMC bioinformatics\/}~{\em 24\/}(1), 434.

\bibitem[\protect\citeauthoryear{Hou, Lin, Torkamani, Wang, and Liu}{Hou et~al.}{2024}]{hou2024adversarial}
Hou, Z., M.~Lin, M.~Torkamani, S.~Wang, and X.~Liu (2024).
\newblock Adversarial robustness in graph neural networks: Recent advances and new frontier.
\newblock In {\em 2024 IEEE 11th International Conference on Data Science and Advanced Analytics (DSAA)}, pp.\  1--2. IEEE.

\bibitem[\protect\citeauthoryear{Kraut and Levine}{Kraut and Levine}{1991}]{kraut1991spatial}
Kraut, R. and M.~Levine (1991).
\newblock Spatial regulation of the gap gene giant during drosophila development.
\newblock {\em Development\/}~{\em 111\/}(2), 601--609.

\bibitem[\protect\citeauthoryear{Levine}{Levine}{2010}]{levine2010transcriptional}
Levine, M. (2010).
\newblock Transcriptional enhancers in animal development and evolution.
\newblock {\em Current Biology\/}~{\em 20\/}(17), R754--R763.

\bibitem[\protect\citeauthoryear{Li, Zhong, and Zhu}{Li et~al.}{2012}]{li2012feature}
Li, R., W.~Zhong, and L.~Zhu (2012).
\newblock Feature screening via distance correlation learning.
\newblock {\em Journal of the American Statistical Association\/}~{\em 107\/}(499), 1129--1139.

\bibitem[\protect\citeauthoryear{Li, MacArthur, Bourgon, Nix, Pollard, Iyer, Hechmer, Simirenko, Stapleton, Hendriks, et~al.}{Li et~al.}{2008}]{li2008transcription}
Li, X.-y., S.~MacArthur, R.~Bourgon, D.~Nix, D.~A. Pollard, V.~N. Iyer, A.~Hechmer, L.~Simirenko, M.~Stapleton, C.~L.~L. Hendriks, et~al. (2008).
\newblock Transcription factors bind thousands of active and inactive regions in the drosophila blastoderm.
\newblock {\em PLoS biology\/}~{\em 6\/}(2), e27.

\bibitem[\protect\citeauthoryear{Liang, Nien, Liu, Metzstein, Kirov, and Rushlow}{Liang et~al.}{2008}]{liang2008zinc}
Liang, H.-L., C.-Y. Nien, H.-Y. Liu, M.~M. Metzstein, N.~Kirov, and C.~Rushlow (2008).
\newblock The zinc-finger protein zelda is a key activator of the early zygotic genome in drosophila.
\newblock {\em Nature\/}~{\em 456\/}(7220), 400.

\bibitem[\protect\citeauthoryear{Lin, Wang, Liu, and Qiu}{Lin et~al.}{2022}]{lin2022survey}
Lin, T., Y.~Wang, X.~Liu, and X.~Qiu (2022).
\newblock A survey of transformers.
\newblock {\em AI open\/}~{\em 3}, 111--132.

\bibitem[\protect\citeauthoryear{Lundberg}{Lundberg}{2017}]{lundberg2017unified}
Lundberg, S. (2017).
\newblock A unified approach to interpreting model predictions.
\newblock {\em arXiv preprint arXiv:1705.07874\/}.

\bibitem[\protect\citeauthoryear{Ma and Fergus}{Ma and Fergus}{2013}]{ma2013visualizing}
Ma, M. and R.~Fergus (2013).
\newblock Visualizing and understanding convolutional networks.
\newblock {\em Computer Vision--ECCV 2014\/}.

\bibitem[\protect\citeauthoryear{Markstein, Zinzen, Markstein, Yee, Erives, Stathopoulos, and Levine}{Markstein et~al.}{2004}]{markstein2004regulatory}
Markstein, M., R.~Zinzen, P.~Markstein, K.-P. Yee, A.~Erives, A.~Stathopoulos, and M.~Levine (2004).
\newblock A regulatory code for neurogenic gene expression in the drosophila embryo.
\newblock {\em Development\/}~{\em 131\/}(10), 2387--2394.

\bibitem[\protect\citeauthoryear{Mor{\'a}n and Jim{\'e}nez}{Mor{\'a}n and Jim{\'e}nez}{2006}]{moran2006tailless}
Mor{\'a}n, {\'E}. and G.~Jim{\'e}nez (2006).
\newblock The tailless nuclear receptor acts as a dedicated repressor in the early drosophila embryo.
\newblock {\em Molecular and Cellular Biology\/}~{\em 26\/}(9), 3446--3454.

\bibitem[\protect\citeauthoryear{Nguyen and Xu}{Nguyen and Xu}{1998}]{nguyen1998drosophila}
Nguyen, H.~T. and X.~Xu (1998).
\newblock Drosophila mef2expression during mesoderm development is controlled by a complex array ofcis-acting regulatory modules.
\newblock {\em Developmental biology\/}~{\em 204\/}(2), 550--566.

\bibitem[\protect\citeauthoryear{Nien, Liang, Butcher, Sun, Fu, Gocha, Kirov, Manak, and Rushlow}{Nien et~al.}{2011}]{nien2011temporal}
Nien, C.-Y., H.-L. Liang, S.~Butcher, Y.~Sun, S.~Fu, T.~Gocha, N.~Kirov, J.~R. Manak, and C.~Rushlow (2011).
\newblock Temporal coordination of gene networks by zelda in the early drosophila embryo.
\newblock {\em PLoS genetics\/}~{\em 7\/}(10), e1002339.

\bibitem[\protect\citeauthoryear{Ribeiro, Singh, and Guestrin}{Ribeiro et~al.}{2016}]{ribeiro2016explaining}
Ribeiro, M.~T., S.~Singh, and C.~Guestrin (2016).
\newblock Explaining the predictions of any classifier.
\newblock In {\em Proceedings of the 22nd ACM SIGKDD International Conference on Knowledge Discovery and Data Mining}, pp.\  1135--1144.

\bibitem[\protect\citeauthoryear{Rivera-Pomar and Jackle}{Rivera-Pomar and Jackle}{1996}]{rivera1996gradients}
Rivera-Pomar, R. and H.~Jackle (1996).
\newblock From gradients to stripes in drosophila embryogenesis: filling in the gaps.
\newblock {\em Trends in Genetics\/}~{\em 12\/}(11), 478--483.

\bibitem[\protect\citeauthoryear{Schulz and Tautz}{Schulz and Tautz}{1994}]{schulz1994autonomous}
Schulz, C. and D.~Tautz (1994).
\newblock Autonomous concentration-dependent activation and repression of kruppel by hunchback in the drosophila embryo.
\newblock {\em Development\/}~{\em 120\/}(10), 3043--3049.

\bibitem[\protect\citeauthoryear{Shah}{Shah}{2016}]{shah2016modelling}
Shah, R.~D. (2016).
\newblock Modelling interactions in high-dimensional data with backtracking.
\newblock {\em Journal of Machine Learning Research\/}~{\em 17\/}(207), 1--31.

\bibitem[\protect\citeauthoryear{Shah and Meinshausen}{Shah and Meinshausen}{2014}]{shah2014random}
Shah, R.~D. and N.~Meinshausen (2014).
\newblock Random intersection trees.
\newblock {\em The Journal of Machine Learning Research\/}~{\em 15\/}(1), 629--654.

\bibitem[\protect\citeauthoryear{Shrikumar, Greenside, and Kundaje}{Shrikumar et~al.}{2017}]{shrikumar2017learning}
Shrikumar, A., P.~Greenside, and A.~Kundaje (2017).
\newblock Learning important features through propagating activation differences.
\newblock In {\em International conference on machine learning}, pp.\  3145--3153. PMlR.

\bibitem[\protect\citeauthoryear{Simon, Friedman, Hastie, and Tibshirani}{Simon et~al.}{2013}]{simon2013sparse}
Simon, N., J.~Friedman, T.~Hastie, and R.~Tibshirani (2013).
\newblock A sparse-group lasso.
\newblock {\em Journal of computational and graphical statistics\/}~{\em 22\/}(2), 231--245.

\bibitem[\protect\citeauthoryear{Stathopoulos, Van~Drenth, Erives, Markstein, and Levine}{Stathopoulos et~al.}{2002}]{stathopoulos2002whole}
Stathopoulos, A., M.~Van~Drenth, A.~Erives, M.~Markstein, and M.~Levine (2002).
\newblock Whole-genome analysis of dorsal-ventral patterning in the drosophila embryo.
\newblock {\em Cell\/}~{\em 111\/}(5), 687--701.

\bibitem[\protect\citeauthoryear{Struhl, Johnston, and Lawrence}{Struhl et~al.}{1992}]{struhl1992control}
Struhl, G., P.~Johnston, and P.~A. Lawrence (1992).
\newblock Control of drosophila body pattern by the hunchback morphogen gradient.
\newblock {\em Cell\/}~{\em 69\/}(2), 237--249.

\bibitem[\protect\citeauthoryear{Sze, Chen, Yang, and Emer}{Sze et~al.}{2017}]{sze2017efficient}
Sze, V., Y.-H. Chen, T.-J. Yang, and J.~S. Emer (2017).
\newblock Efficient processing of deep neural networks: A tutorial and survey.
\newblock {\em Proceedings of the IEEE\/}~{\em 105\/}(12), 2295--2329.

\bibitem[\protect\citeauthoryear{Tibshirani}{Tibshirani}{1996}]{tibshirani1996regression}
Tibshirani, R. (1996).
\newblock Regression shrinkage and selection via the lasso.
\newblock {\em Journal of the Royal Statistical Society Series B: Statistical Methodology\/}~{\em 58\/}(1), 267--288.

\bibitem[\protect\citeauthoryear{Xu, Ma, Liu, Deb, Liu, Tang, and Jain}{Xu et~al.}{2020}]{xu2020adversarial}
Xu, H., Y.~Ma, H.-C. Liu, D.~Deb, H.~Liu, J.-L. Tang, and A.~K. Jain (2020).
\newblock Adversarial attacks and defenses in images, graphs and text: A review.
\newblock {\em International journal of automation and computing\/}~{\em 17}, 151--178.

\bibitem[\protect\citeauthoryear{Yang, Zhang, Li, and Huang}{Yang et~al.}{2019}]{yang2019feature}
Yang, G., L.~Zhang, R.~Li, and Y.~Huang (2019).
\newblock Feature screening in ultrahigh-dimensional varying-coefficient cox model.
\newblock {\em Journal of Multivariate Analysis\/}~{\em 171\/}(C), 284--297.

\bibitem[\protect\citeauthoryear{Yuan and Lin}{Yuan and Lin}{2006}]{yuan2006model}
Yuan, M. and Y.~Lin (2006).
\newblock Model selection and estimation in regression with grouped variables.
\newblock {\em Journal of the Royal Statistical Society Series B: Statistical Methodology\/}~{\em 68\/}(1), 49--67.

\bibitem[\protect\citeauthoryear{Zeitlinger, Zinzen, Stark, Kellis, Zhang, Young, and Levine}{Zeitlinger et~al.}{2007}]{zeitlinger2007whole}
Zeitlinger, J., R.~P. Zinzen, A.~Stark, M.~Kellis, H.~Zhang, R.~A. Young, and M.~Levine (2007).
\newblock Whole-genome chip--chip analysis of dorsal, twist, and snail suggests integration of diverse patterning processes in the drosophila embryo.
\newblock {\em Genes \& development\/}~{\em 21\/}(4), 385--390.

\bibitem[\protect\citeauthoryear{Zhao, Dong, Luo, Wu, Bu, Qi, Luo, and Zhao}{Zhao et~al.}{2021}]{zhao2021deepomix}
Zhao, L., Q.~Dong, C.~Luo, Y.~Wu, D.~Bu, X.~Qi, Y.~Luo, and Y.~Zhao (2021).
\newblock Deepomix: a scalable and interpretable multi-omics deep learning framework and application in cancer survival analysis.
\newblock {\em Computational and structural biotechnology journal\/}~{\em 19}, 2719--2725.

\bibitem[\protect\citeauthoryear{Zhou and Troyanskaya}{Zhou and Troyanskaya}{2015}]{zhou2015predicting}
Zhou, J. and O.~G. Troyanskaya (2015).
\newblock Predicting effects of noncoding variants with deep learning--based sequence model.
\newblock {\em Nature methods\/}~{\em 12\/}(10), 931--934.

\bibitem[\protect\citeauthoryear{Zintgraf, Cohen, Adel, and Welling}{Zintgraf et~al.}{2017}]{zintgraf2017visualizing}
Zintgraf, L.~M., T.~S. Cohen, T.~Adel, and M.~Welling (2017).
\newblock Visualizing deep neural network decisions: Prediction difference analysis.
\newblock {\em arXiv preprint arXiv:1702.04595\/}.

\bibitem[\protect\citeauthoryear{Zou and Hastie}{Zou and Hastie}{2005}]{zou2005regularization}
Zou, H. and T.~Hastie (2005).
\newblock Regularization and variable selection via the elastic net.
\newblock {\em Journal of the Royal Statistical Society Series B: Statistical Methodology\/}~{\em 67\/}(2), 301--320.

\bibitem[\protect\citeauthoryear{Zou and Li}{Zou and Li}{2008}]{zou2008one}
Zou, H. and R.~Li (2008).
\newblock One-step sparse estimates in nonconcave penalized likelihood models.
\newblock {\em Annals of statistics\/}~{\em 36\/}(4), 1509--1533.

\end{thebibliography}
